\documentclass[10pt,journal,compsoc]{IEEEtran}
\ifCLASSOPTIONcompsoc
  \usepackage[nocompress]{cite}
\else
  \usepackage{cite}
\fi

\usepackage{times}
\usepackage{epsfig}
\usepackage{graphicx}
\usepackage{amsmath}
\usepackage[caption=false,font=footnotesize]{subfig}
\usepackage{threeparttable}
\usepackage{amssymb}
\usepackage{tabularx}
\usepackage{booktabs}
\usepackage{multirow}
\usepackage{multicol}
\usepackage{footmisc}
\usepackage{csquotes}
\usepackage{rotating}
\usepackage{caption}
\usepackage{float}
\usepackage{array, makecell}
\usepackage{threeparttable}
\usepackage{tablefootnote}
\usepackage{booktabs}
\usepackage[export]{adjustbox}
\usepackage{array}
\usepackage{fancyhdr}
\newcolumntype{P}[1]{>{\centering\arraybackslash}p{#1}}
\newcolumntype{M}[1]{>{\centering\arraybackslash}m{#1}}
\usepackage{dblfloatfix}
\usepackage{amsmath}

\usepackage[pagebackref=true,breaklinks=true,letterpaper=true,colorlinks,bookmarks=false]{hyperref}

\hypersetup{
  colorlinks, linkcolor=red
}
\makeatletter
\g@addto@macro{\UrlBreaks}{\UrlOrds}
\makeatother

\usepackage{capt-of,etoolbox}

\renewcommand{\mkbegdispquote}[2]{\itshape}

\begin{document}
%
\title{Infant-ID: Fingerprints for Global Good}
%
%
%
%

\author{Joshua J. Engelsma,~Student Member,~IEEE,
        Debayan Deb,~Student Member,~IEEE,
        Kai Cao,
        Anjoo Bhatnagar,
        Prem S. Sudhish~Member,~IEEE,
        Anil K. Jain,~Life~Fellow,~IEEE
\IEEEcompsocitemizethanks{\IEEEcompsocthanksitem J. Engelsma, D. Deb and A. K. Jain are with the
Department of Computer Science and Engineering, Michigan State University, East Lansing, MI, 48824 USA. E-mail: {engelsm7, debdebay, jain}@cse.msu.edu\protect\\
\IEEEcompsocthanksitem K. Cao is a Senior Researcher at Goodix, San Diego, CA. Email: caokai0505@gmail.com\protect\\

\IEEEcompsocthanksitem A. Bhatnagar is with the Saran Ashram hospital, Dayalbagh, UP, India
282005. Email: dranjoo@gmail.com\protect\\

\IEEEcompsocthanksitem P.S. Sudhish is with the Dayalbagh Educational Institute, Dayalbagh, UP, India
282005. Email: pss@alumni.stanford.edu

}
}

%
%

\markboth{}%
{Engelsma \MakeLowercase{\textit{et al.}}: Infant-ID: Fingerprints for Global Good}
%



\twocolumn[{%
\renewcommand\twocolumn[1][]{#1}%
\begin{@twocolumnfalse}
    \maketitle
  \end{@twocolumnfalse}
\begin{center}
\footnotesize
    \centering
    \begin{minipage}{0.16\linewidth}
    \includegraphics[width=\linewidth]{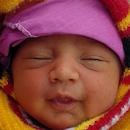}
    \end{minipage}
    \begin{minipage}{0.16\linewidth}
    \includegraphics[width=\linewidth]{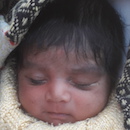}
    \end{minipage}
    \begin{minipage}{0.16\linewidth}
    \includegraphics[width=\linewidth]{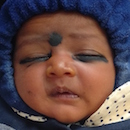}
    \end{minipage}
    \begin{minipage}{0.16\linewidth}
    \includegraphics[width=\linewidth]{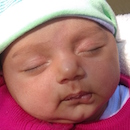}
    \end{minipage}
    \begin{minipage}{0.16\linewidth}
    \includegraphics[width=\linewidth]{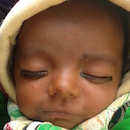}
    \end{minipage}
    \begin{minipage}{0.16\linewidth}
    \includegraphics[width=\linewidth]{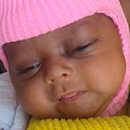}
    \end{minipage}\\ \vspace{0.3em}
    \begin{minipage}{0.16\linewidth}
    \includegraphics[width=\linewidth]{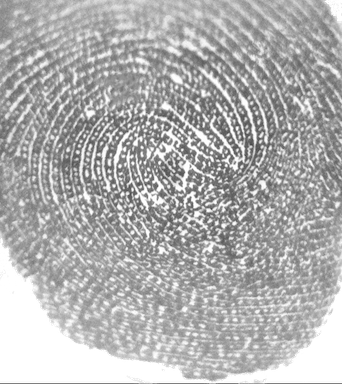}
    \centering {1 week old}
    \end{minipage}
    \begin{minipage}{0.16\linewidth}
    \includegraphics[width=\linewidth]{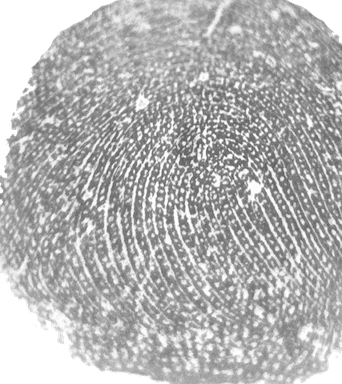}
    \centering {2 weeks old}
    \end{minipage}
    \begin{minipage}{0.16\linewidth}
    \includegraphics[width=\linewidth]{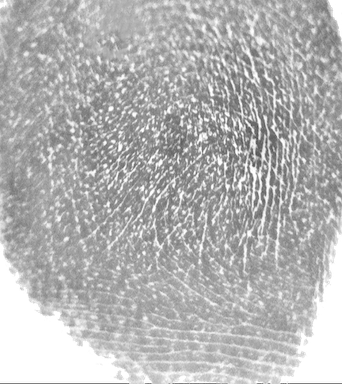}
    \centering {4 weeks old}
    \end{minipage}
    \begin{minipage}{0.16\linewidth}
    \includegraphics[width=\linewidth]{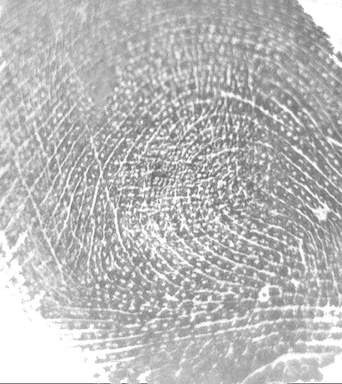}
    \centering {6 weeks old}
    \end{minipage}
    \begin{minipage}{0.16\linewidth}
    \includegraphics[width=\linewidth]{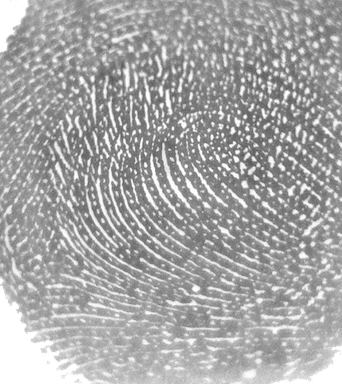}
    \centering {8 weeks old}
    \end{minipage}
    \begin{minipage}{0.16\linewidth}
    \includegraphics[width=\linewidth]{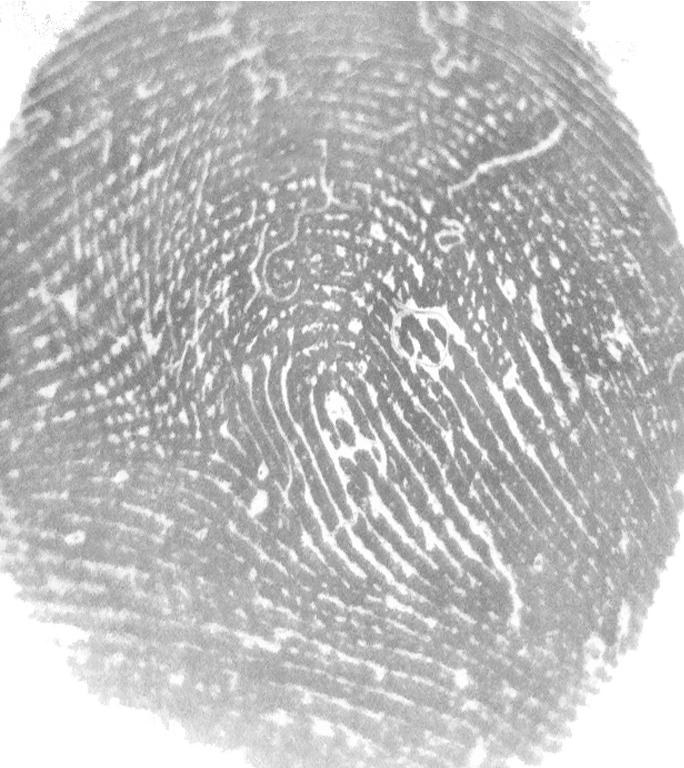}
    \label{fig:paper_cut}
    \centering {12 weeks old}
    \end{minipage}\\
    \begin{minipage}{\linewidth}
    \begin{figure}[H]
    \captionof{figure}{Face images (top row) and corresponding left thumb fingerprints (bottom row) of six different infants under 3 months of age. Face images were captured by a \emph{Xiaomi MI A1} smartphone camera and fingerprint images were captured by the 1,900 ppi RaspiReader designed by Engelsma \textit{et al.}~\cite{engelsma1, infant_prints} at the Saran Ashram Hospital, a charitable organization in Dayalbagh, Agra, India.}
    \label{fig:spoof_types}
    \end{figure}
    \end{minipage}
\end{center}%
}]


{
  \renewcommand{\thefootnote}%
    {\fnsymbol{footnote}}
  \footnotetext{
  J.J. Engelsma, D. Deb and A. K. Jain are with the Department of Computer Science and Engineering, Michigan State University, East Lansing, MI, 48824. E-mail: \{engelsm7, debdebay, jain\}@cse.msu.edu \\
  \indent\indent Kai Cao is a Senior Researcher at Goodix in San Diego, CA. Email: caokai0505@gmail.com \\
  \indent\indent A. Bhatnagar is with the Saran Ashram hospital, Dayalbagh, UP, India 282005. Email: dranjoo@gmail.com \\
  \indent\indent P.S. Sudhish is with the Dayalbagh Educational Institute, Dayalbagh, UP, India 282005. Email: pss@alumni.stanford.edu
  }
}

\begin{abstract}
In many of the least developed and developing countries, a multitude of
infants continue to suffer and die from vaccine-preventable
diseases and malnutrition. Lamentably, the lack of official identification documentation makes it exceedingly difficult to track which infants have been vaccinated and which infants have received nutritional supplements. Answering these questions could prevent this infant suffering and premature death around the world. To that end, we propose Infant-Prints, an end-to-end, low-cost, infant fingerprint recognition system. Infant-Prints is comprised of our (i) custom built, compact, low-cost (85 USD), high-resolution
(1,900 ppi), ergonomic fingerprint reader, and (ii) high-resolution infant fingerprint matcher. To evaluate the efficacy of Infant-Prints, we collected a longitudinal infant fingerprint database captured in 4 different sessions over a 12-month time span (December 2018 to January 2020), from 315 infants at the Saran Ashram Hospital, a charitable hospital in Dayalbagh, Agra, India. Our experimental results demonstrate, for the first time, that Infant-Prints can deliver accurate
and reliable recognition (over time) of infants enrolled between the ages of 2-3
months, in time for effective delivery of vaccinations, healthcare, and nutritional supplements (\textbf{TAR=95.2\% @ FAR = 1.0\%} for
infants aged 8-16 weeks at enrollment and authenticated 3 months later)\footnote{A preliminary version of this paper was present at CVPRW Computer Vision for Global Challenges, Long Beach, CA, 2019.}.
\end{abstract}





%
\IEEEpeerreviewmaketitle

\section{Introduction}\label{sec:introduction}

%
%
%
%
\IEEEPARstart{I}{n} an effort to develop a better world for people and our planet over the next decade, the United Nations (UN) has set forth 17 Sustainable Development Goals. Perhaps the most imperative of these goals is Goal \#3: 

\begin{quote}
\textit{``Ensuring healthy lives and promoting well-being for all, at all ages.''~\cite{sdg}}
\end{quote} 

\begin{figure*}[!t]
    \centering
    \captionsetup[subfigure]{labelformat=empty}
    \subfloat[]{\includegraphics[height=1.8in]{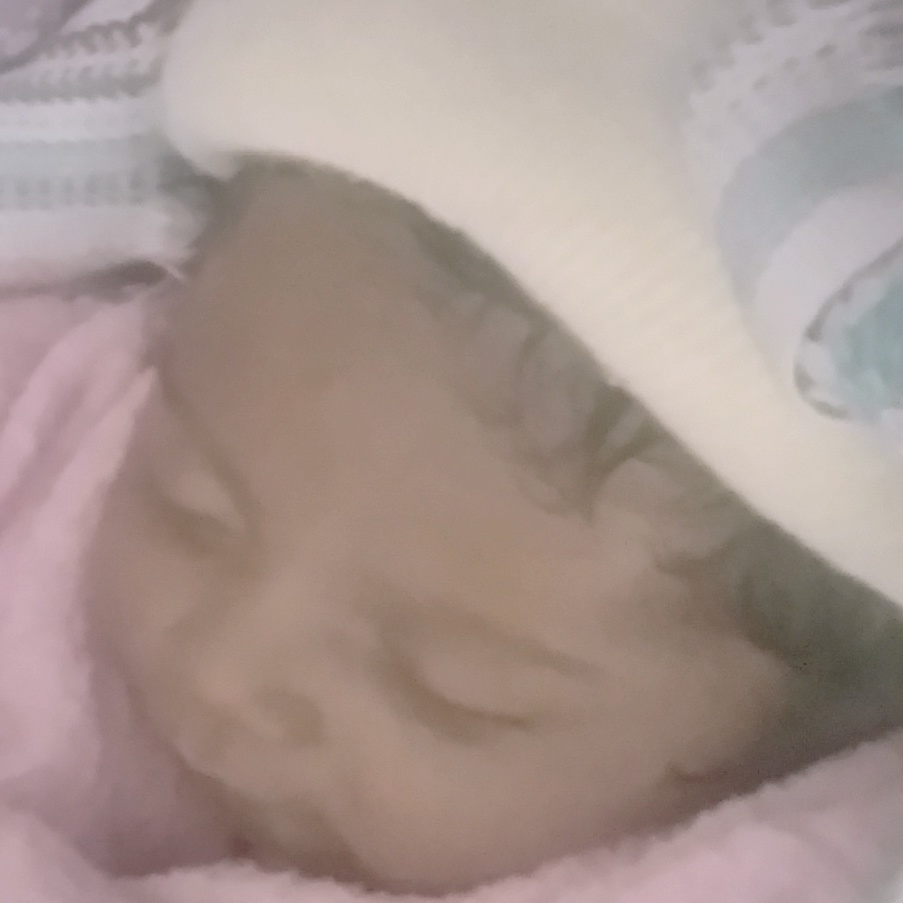}}\hfill
    \subfloat[]{\includegraphics[height=1.8in]{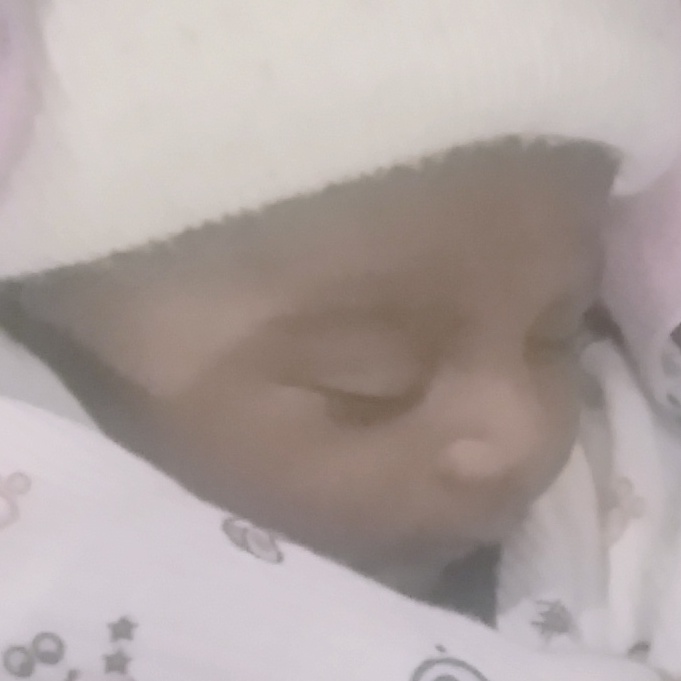}}\hfill
    \subfloat[]{\includegraphics[height=1.8in]{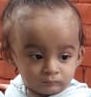}}\hfill
    \subfloat[]{\includegraphics[height=1.8in]{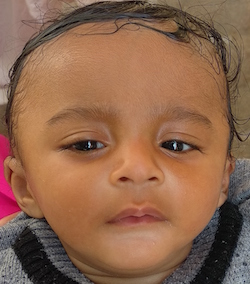}}\\
    \subfloat[3 months]{\includegraphics[height=1.8in]{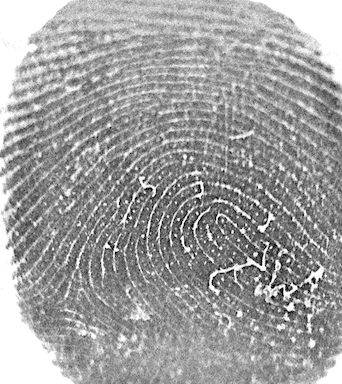}}\hfill
    \subfloat[3 months, 2 days]{\includegraphics[height=1.8in]{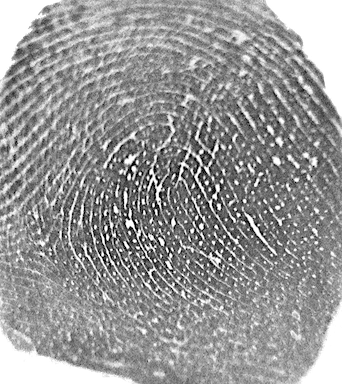}}\hfill
    \subfloat[6 months]{\includegraphics[height=1.8in]{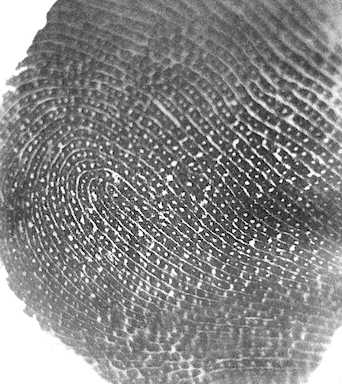}}\hfill
    \subfloat[12 months]{\includegraphics[height=1.8in]{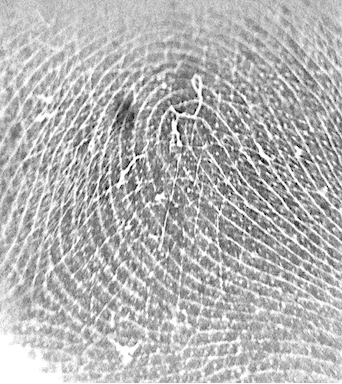}}
    \caption{Face images (top row) and corresponding left thumb fingerprints (bottom row) of an infant, \emph{Meena Kumari}, acquired on (a) December 16, 2018 (Meena was 3 months old), (b) December 18, 2018 (3 months, 2 days old),  (c) March 5, 2019 (6 months old), and (d) September 17, 2019 (12 months old) at Saran Ashram Hospital, Dayalbagh, India. Note that as Meena ages, fingerprint details emerge such as visible pores. This level of detail is enabled by our 1,900 ppi reader.}
    \label{fig:frontpage}
\end{figure*}

\noindent The UN has also specified that the mortality rate of children under 5 is a key indicator of the progress made in achieving this goal\footnote{\url{https://data.unicef.org/topic/child-survival/under-five-mortality/}}. Although the child mortality rate has been shrinking over the past two decades (from 93 deaths in 1990 to 39 deaths in 2019 per 1000 live births), an estimated 5.3 million children under the age of five died in 2018 globally. Furthermore, there are currently more than 600 million children living worldwide between the ages of 0-5 years~\cite{age_structure}, with over 353,000 more newborns setting foot on the planet each and every day~\cite{birth_rate}. A majority of these births take place in the poorest regions of the world, where it is likely that neither the infants nor their parents will have access to any official identification documents\footnote{Selecting and assigning a name to the newborns can be a drawn out process in developing countries in which parents consult immediate family members or even an astrologer for a
proper name. While deciding upon a name, the infant is simply referred to as ``baby" or
``daughter of", or ``son of".}. Even if the infant has an official ID, it may be fraudulent or shared with others~\cite{fraud, fraud2, fraud3}. Without legitimate and verifiable identification, infants are often denied access to healthcare, immunization, and nutritional supplements. This is especially problematic for infants\footnote{Infants are considered to be in the 0-12 months age range, whereas, toddlers and preschoolers are within 1-3 and 3-5 years of age, respectively~\cite{age_classification}.} (newborns to 12 months), given that they are at their most critical stage of development.  

\begin{table*}[!t]
\centering
\caption{Related work on child fingerprint recognition.}
\label{tab:related}
\begin{threeparttable}
\resizebox{\linewidth}{!}{
\begin{tabular}{lcccccl}
\noalign{\hrule height 1.5pt}
\textbf{Study} & \textbf{Year}  & \textbf{Fingerprint Resolution} & \textbf{\# Subjects} &\textbf{Age at Enrolment} &\textbf{Time Lapse} &\textbf{Findings} \\
\noalign{\hrule height 1pt}
Galton~\cite{galton} & 1899 & Inked Impressions & 1 & 0 year & 0 - 4.5 years & Recognition is feasible for children over 2.5 years\\
\hline
TNO~\cite{tno} & 2005 & 500 ppi & 161 & 0 - 13 years & N/A\tnote{*} & Recognition is challenging for children below 4 years\\
\hline
BIODEV II~\cite{biodev} & 2007 & 500 ppi & 300 & 0 - 12 years & N/A\tnote{*} & Difficult to capture fingerprints for children $<$ 12 years\\
\hline
UltraScan~\cite{ultrascan} & 2006-2009  & 500 ppi & 308 & 0 - 18 years & 3 years & No insight for children below 5 years\\
\hline
Aadhar~\cite{aadhar} & 2009 & 500ppi & 1.25B & Enrolled at 5 years of age & N/A & Recognition of children under 5 years of age is challenging\\& & & & Re-enrolled at 15 years of age & &\\
\hline
JRC~\cite{jrc} & 2013 & 500 ppi & 2611 & 0 - 12 years & 2 - 4 years & Recognition of children under 6 years of age is difficult\\
\hline
Jain~\textit{et al.}~\cite{jain} & 2016 & 1,270 ppi & 309 & 0 - 5 years & 1 year & Feasible to recognize children over 6 months\\
\hline
Saggese~\textit{et al.}~\cite{gates} & 2019 & 3,400 ppi & 142 & 0 - 6 months & variable length & High authentication accuracy (TAR = 85\%-96\% @ FAR = 0.1\%),\\& & & & & & but unknown time lapse between enrollment and authentication\tnote{1}.\\
\hline
Infant-Prints~\cite{infant_prints} & 2019 & 1,900 ppi & 194 & 0 - 3 mos. & 3 mos. &  TAR = 66.7\%,  75.4\%, and 90.2\% @ FAR = 0.1\% for infants enrolled\\ & & & & & & at ages [0-3 months], [1-3 months], and  [2-3 months], respectively.\\
\hline
Preciozzi~\textit{et al.}~\cite{infant_prints} & 2020 & 500 ppi & 16,865 & 0 - 18 years & 10 years &  TAR = 1.25\%,  7.57\%, and 15.61\% @ FAR = 0.1\% for infants enrolled\\ & & & & & & at ages [0-1 month], [1-2 months], and  [2-3 months], respectively.\\
\hline
This study & 2020 & 1,900 ppi & 315 & 0 - 3 months & 1 year & TAR = 92.8\% @ FAR = 0.1\% for infants enrolled\\ & & & & & & at age of 2-3 months, respectively.\\
\noalign{\hrule height 1.5pt}
\end{tabular}
}%
\begin{tablenotes}\footnotesize
\item[*] No time span available for these studies.
\item[1] Scores from across all time lapses (weeks or months) are aggregated when computing the fingerprint recognition error rates. \\This inflates the true longitudinal recognition performance.
\end{tablenotes}
\end{threeparttable}
\end{table*}

The downstream problems caused by lack of proper infant ID in the planet's least-developed countries can be quantitatively seen in the flat lining of global vaccination coverage. In particular, from 2015 to 2018, the percentage of children who have received their full course of three-dose diphtheria-tetanus-pertussis (DTP3) routine immunizations remains at about 85\%~\cite{global_immunization}.  This falls short of the GAVI Alliance (formerly Global Alliance for Vaccines and Immunization\footnote{\url{https://bit.ly/1i7s8s2}}) target of achieving global immunization coverage of 90\% by 2020.  According to UNICEF, 25 million children do not receive
proper annual vaccination, leading to 1.5 million child deaths per annum from vaccine-preventable diseases\footnote{\url{https://www.unicef.org/immunization}}. The World Health Organization (WHO) suggests that inadequate monitoring and supervision and lack of official identification documents (making it exceedingly difficult to accurately track vaccination schedules) are key factors\footnote{\url{https://bit.ly/1pWn6Gn}}.

Infant identification is also urgently needed to effectively provide nutritional supplements. The World Food Program (WFP), a leading humanitarian organization fighting hunger worldwide, assists close to 100 million people in some of the poorest regions of the world\footnote{\url{https://evaw-un-inventory.unwomen.org/fr/agencies/wfp}}. 
However, often the food never reaches the intended beneficiaries because of fraud in the distribution system~\cite{fraud, fraud2, fraud3}. For example, the WFP found that in Yemen, a country with 12 million starving residents, food distribution records are falsified and relief is being given to people not entitled to it, preventing those who actually need aid from receiving it~\cite{fraud2,fraud3}. 

Accurate and reliable infant recognition would also assist in baby swapping prevention\footnote{\url{https://bit.ly/2U5eAHn}}, identifying missing or abducted children, and access to government benefits, healthcare, and financial services throughout an infant's lifetime. 

While many different countries around the world could benefit from accurate, fast, and user-friendly infant recognition and tracking methods for the aforementioned reasons, the benefits would be most palpable in the least developed and developing countries which lack robust systems for widespread identification coverage. Notably, the development level of a country is highly correlated with the country's identification coverage. In particular, 36\% of the population in low-income economies lack
official IDs, compared to 22\% and 9\% in lower-middle and
upper-middle income economies~\cite{global_identity}, respectively. Unfortunately, many of those missing official identification in these countries are children under the age of 18. This lack of official identification amongst children has prompted the World Bank Group to adopt core Principles
of Identification, which include \emph{universal coverage for all from birth to death, without discrimination}~\cite{principles}. Following this central tenant from the World Bank Group's Principles for Identification via a robust infant-recognition system would significantly alleviate infant suffering around the world, particularly in least developed, and developing countries.
 
Our problem description can be concisely stated as follows. Given an  infant who comes to a vaccination clinic or a food distribution center, we would like to answer the following two questions: 

\begin{itemize}
\item \emph{``Have we seen this child before?"}
\item \emph{``Is this the child who their parents claim them to be?"}
\end{itemize}

As we show in the next section, biometric recognition~\cite{handbook} is the only way to accurately and reliably answer these two questions. While biometrics is now a mature field and billions of teenagers and adults have been using it to authenticate themselves, children, particularly infants and toddlers, cannot yet utilize biometrics to get a unique and verifiable digital identity.

\section{Biometrics for Infants}
Conventional identification documents (paper records) are impractical for infant recognition in many of the least developed and developing countries because they are not securely linked to a specific infant. Furthermore, they may be fraudulent~\cite{fraud}, lost, or stolen. We posit that a more accurate, robust, and verifiable means of infant recognition is through the use of \emph{biometric recognition}. Biometric recognition, or simply biometrics, is the automatic recognition of a person by their body traits (\textit{e.g.} face, fingerprint, palmprint, iris) via an association of the trait to an identity (\textit{e.g.}, a person's name or other externally assigned identifiers and meta-data)~\cite{handbook}.

Designing a biometric recognition system for infants is a significant challenge. The world's largest successful biometric recognition system, India's \emph{Aadhaar}, utilizes both the irises, all ten fingerprints and face to uniquely identify (de-duplication) and then link a citizen to a 12-digit unique identifier, called Aadhaar (meaning foundation in Hindi). So far, over 1.2 billion Indian residents have been assigned an Aadhar number\footnote{\url{https://bit.ly/2zqrBSq}}. However, Aadhaar starts biometric enrollment at the age of 5 years, leaving children under the age of 5 years without a biometrics-linked Aadhaar ID. The lack of an accurate biometric solution for all children (infants, toddlers and preschoolers) has prompted UNICEF to state that biometrics for children \emph{``requires further extensive research, as there is a critical lack of verifiable performance data on most of the technologies currently in use with children (particularly for longitudinal use over extended periods)"}~\cite{unicef}.

Compounding the difficulty in designing a biometric recognition system for infants is that a majority of the biometric modalities are not appropriate for infants. Note that an infant is not a ``cooperative" subject and cannot understand instructions to stay still, provide a neutral facial expression, open an eye, or open a fist. Furthermore, an infant's face~\cite{face_infants} changes daily due to rapid aging from infanthood to childhood~\cite{uni}. Iris image capture~\cite{iris_infants} is challenging for infants when the child is sleeping or crying. Furthermore, parents may have concerns about the safety of capturing an infant's iris image. Footprint recognition~\cite{footprint, bigtoe}, requires removing socks and shoes and cleaning the infant's feet, and palmprint recognition~\cite{palmprint} requires opening an infant's entire hand where the concavity of the palm makes it difficult to image.

\begin{figure*}[!t]
    \centering
    \includegraphics[width=\linewidth]{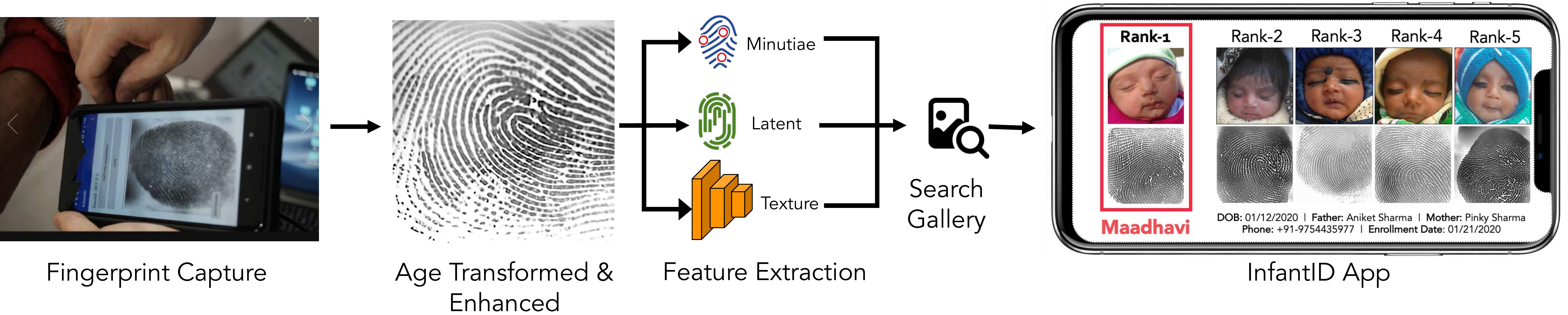}
    \caption{Overview of the Infant-Prints system.}
    \label{fig:ip_overview}
\end{figure*}

\subsection{Fingerprints for Infant-ID}
We posit that fingerprint is the most promising biometric trait for infant recognition for the following several reasons. (i) Uniqueness (individuality) studies on fingerprints have shown that they are indeed different for different fingers even of the same person, including identical twins~\cite{individuality, individuality2}. (ii) Biological evidence suggests fingerprint patterns are present at birth~\cite{birth1,birth2,fingerprint_fetus}. (iii) While the friction ridge patterns on our fingers may grow or fade over time, longitudinal studies on fingerprint recognition for adults~\cite{yoon} and children~\cite{infant_jain} show that the fingerprint recognition accuracy does not change appreciably over time  (approximately 12 year time gap for adults). (iv) Fingerprints are the most convenient and socially acceptable biometric to capture from infants~\cite{jain}.

Fingerprint recognition of infants comes with its own challenges and requirements, including:
\begin{enumerate}
    \item A compact, low-cost, ergonomic, high-resolution (to accommodate small inter-ridge spacings), and high throughput fingerprint reader.
    \item A robust and accurate fingerprint matcher to accommodate low quality (distorted, dirty, wet, dry, motion blurred), high-resolution fingerprint images.
\end{enumerate}
As such, prevailing COTS fingerprint recognition systems, designed primarily for an adult population, are not feasible for infant fingerprint recognition.

The immunization schedules outlined by the Centers for Disease Control and Prevention (CDC) recommend administering the infant with the first dosage of vaccines by the age of 3 months\footnote{\url{https://www.cdc.gov/vaccines/schedules/hcp/imz/child-adolescent.html}}. In addition, infant mortality rates are severely affected by inadequite nutrition between ages 0-6 months\footnote{\url{https://www.ncbi.nlm.nih.gov/books/NBK148967/}}. Therefore, an infant fingerprint recognition system must be able to identify infants enrolled below the age of 3 months with high accuracy. Table~\ref{tab:related} summarizes prior work on infant fingerprint recognition. 
Among these, there are only three studies~\cite{gates, growth2, infant_prints} which investigate the feasibility of recognizing infants under the age of 3 months at enrollment. (i) While the infant fingerprint recognition results reported in~\cite{gates} by Saggese \textit{et al.} seem promising, they aggregate scores from all time lapses (weeks or months) for computing the fingerprint recognition error rates which inflates the true longitudinal recognition performance. (ii) Preciozzi~\textit{et al.} report poor infant recognition results (TAR = $15.61\%$ @ FAR=$0.1\%$ for 2-3 month old age group). (iii) Our preliminary study on infant fingerprint recognition~\cite{infant_prints} utilized a custom 1,900 ppi infant fingerprint reader, however, the matcher was not designed to fully utilize the high-resolution imagery (instead using existing matchers designed for 500 ppi images). Furthermore, the matcher did not incorporate any enhancement or aging of the friction ridge pattern. Finally, our preliminary study was conducted for 194 infants across a maximum time lapse of 3 months. In contrast, the current work includes 315 infants with longitudinal data of up to a one year time lapse. 

We have designed a first of its kind, end to end, high-resolution, infant fingerprint recognition system, called \textbf{Infant-Prints} (Fig.~\ref{fig:ip_overview}). Infant-Prints is comprised of (i) a low-cost, high resolution, infant fingerprint reader, (ii) an infant fingerprint enhancement module, (iii) an aging model to compensate for the growth of the infant's fingerprints over time, (iv) a high resolution minutiae matcher, (v) a high resolution texture matcher, and (vi) a COTS latent matcher. The experimental results of Infant-Prints evaluated on our longitudinal infant dataset indicate that indeed, it is possible to enroll infants at ages younger than 3 months and accurately recognize them months later based only upon their fingerprints \textbf{TAR=95.2\%@FAR=1.0\%},\textbf{TAR=92.8\%@FAR=0.1\%}  (for infants enrolled at 2-3 months of age, and authenticated 3 months later).

\begin{figure*}[!t]
    \centering
    \subfloat[]{\includegraphics[width=0.43\textwidth]{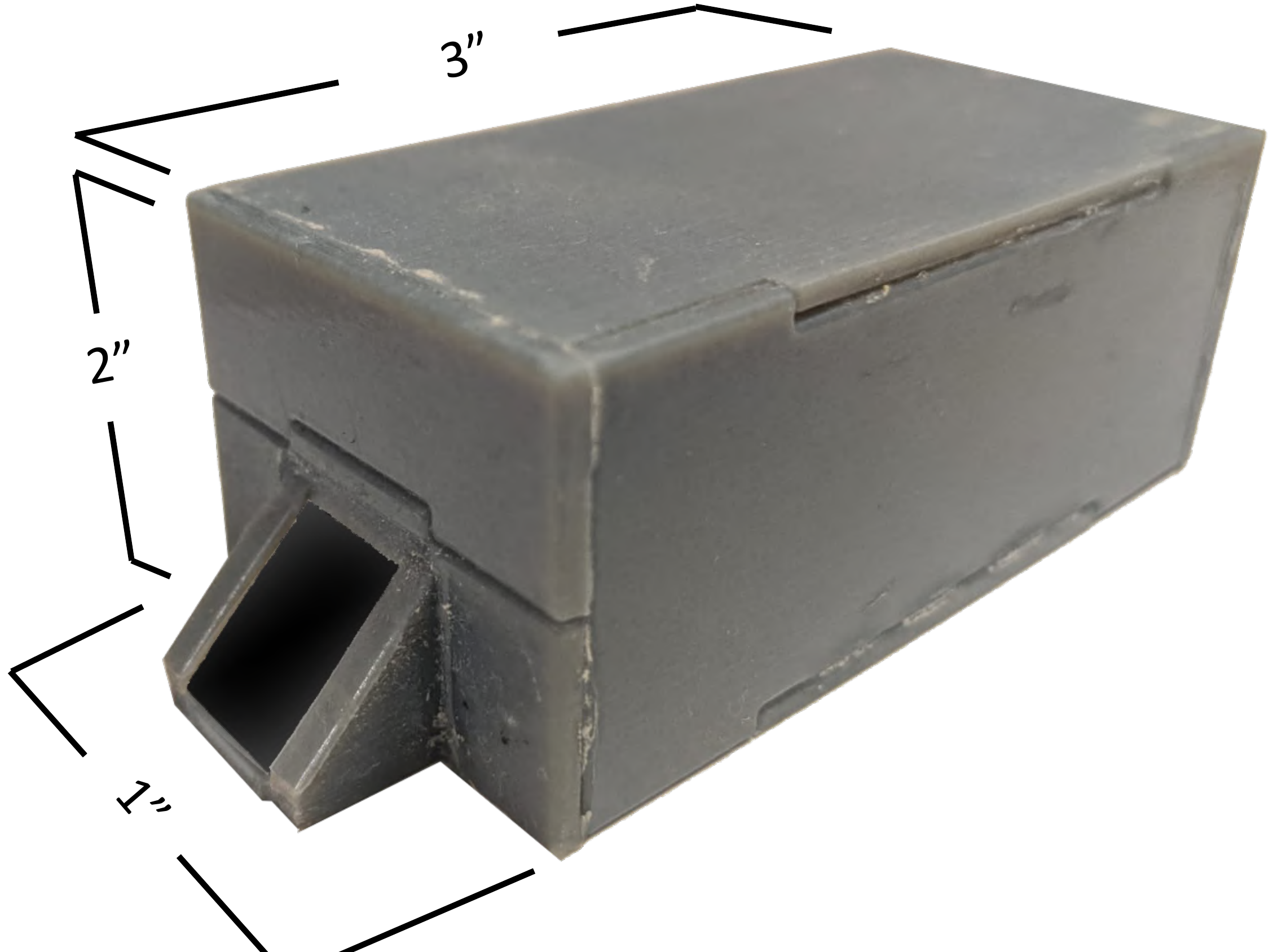}}
    \subfloat[]{\includegraphics[width=0.57\textwidth]{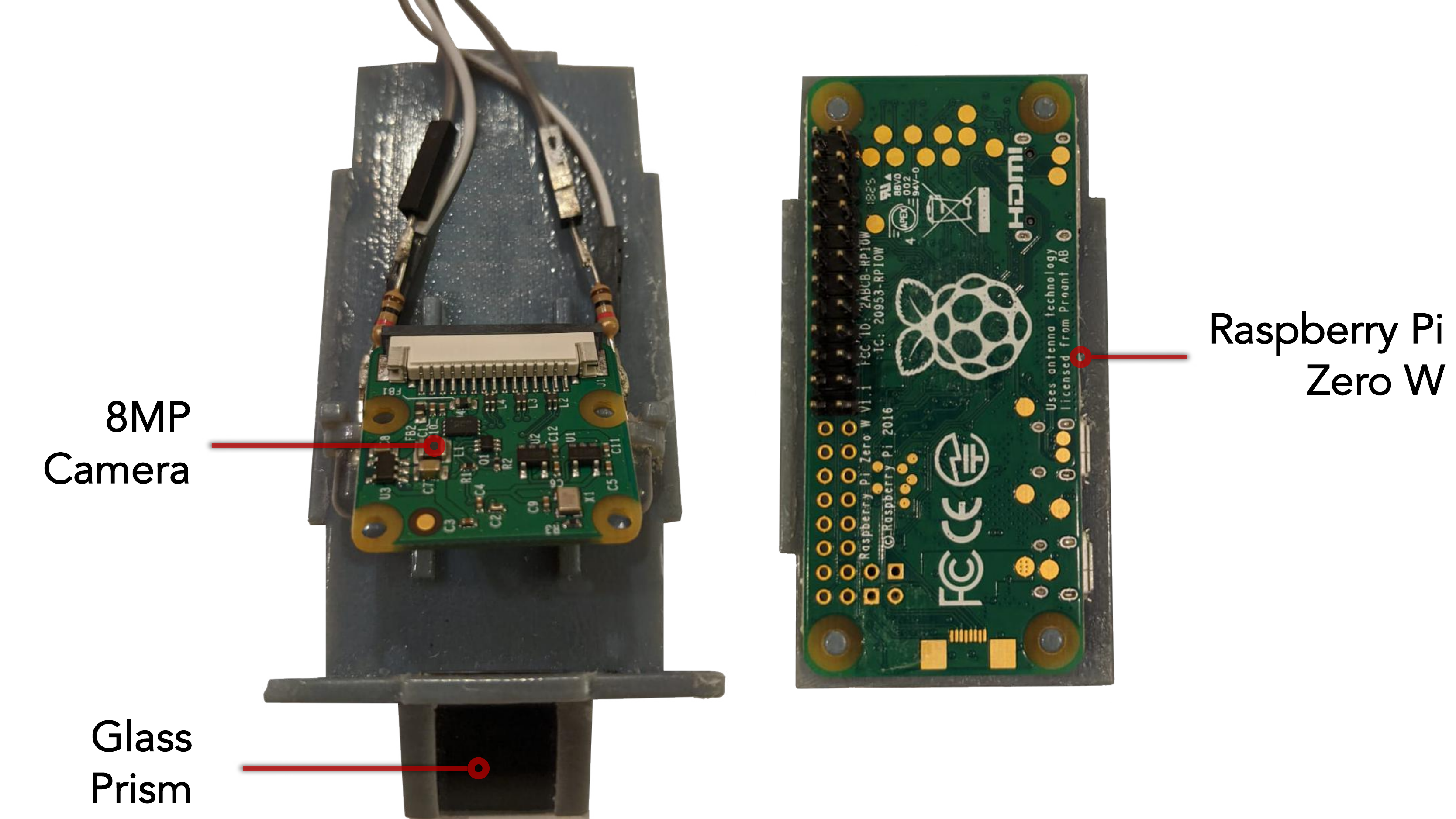}}
    \caption{Prototype of the 1,900 ppi compact (1" x 2" x 3"), ergonomic fingerprint reader. An infant's finger is placed on the glass prism with the operator applying slight pressure on the finger. The capture time is 500 milliseconds. The prototype can be assembled in less than 2 hours. See the video at:~\url{https://www.youtube.com/watch?v=f8tYbE9Cwd0}.}
    \label{fig:reader}
\end{figure*}

Concisely, the contributions of this paper are summarized as follows: 
\begin{itemize}
    \item Design and prototyping of a compact (1"$\times$2"$\times$3"), low-cost (85 USD), high-resolution (1,900 ppi), ergonomic fingerprint reader for infants (Fig.~\ref{fig:reader}). This reader is much smaller and better designed for infants than our earlier open sourced fingerprint reader proposed in~\cite{engelsma1}. We also prototype a contactless version of our fingerprint reader (Fig.~\ref{fig:contactless_reader}) in order to compare contact-based sensing technologies with contactless sensing technologies when used for infants. 
    \item Collection of a longitudinal infant fingerprint database comprised of 315 infants (0-3 months) over 4 separate sessions separated by 13 months (between December 2018 and January 2020). The data was collected at the Saran Ashram hospital, Dayalbagh, India. 
    \item A high resolution, fingerprint matcher for infants which incorporates infant fingerprint aging and enhancement modules together with high resolution texture and minutiae matchers. 
    \item Demonstrated accurate recognition accuracy of TAR=95.2\% (92.8\%) @ FAR=1.0\% (0.1\%) for infants enrolled at 2-3 months, and authenticated 3 months later.
\end{itemize}

\begin{figure}[!t]
    \centering
    \subfloat[500 ppi commercial reader]{\includegraphics[height=1.21in]{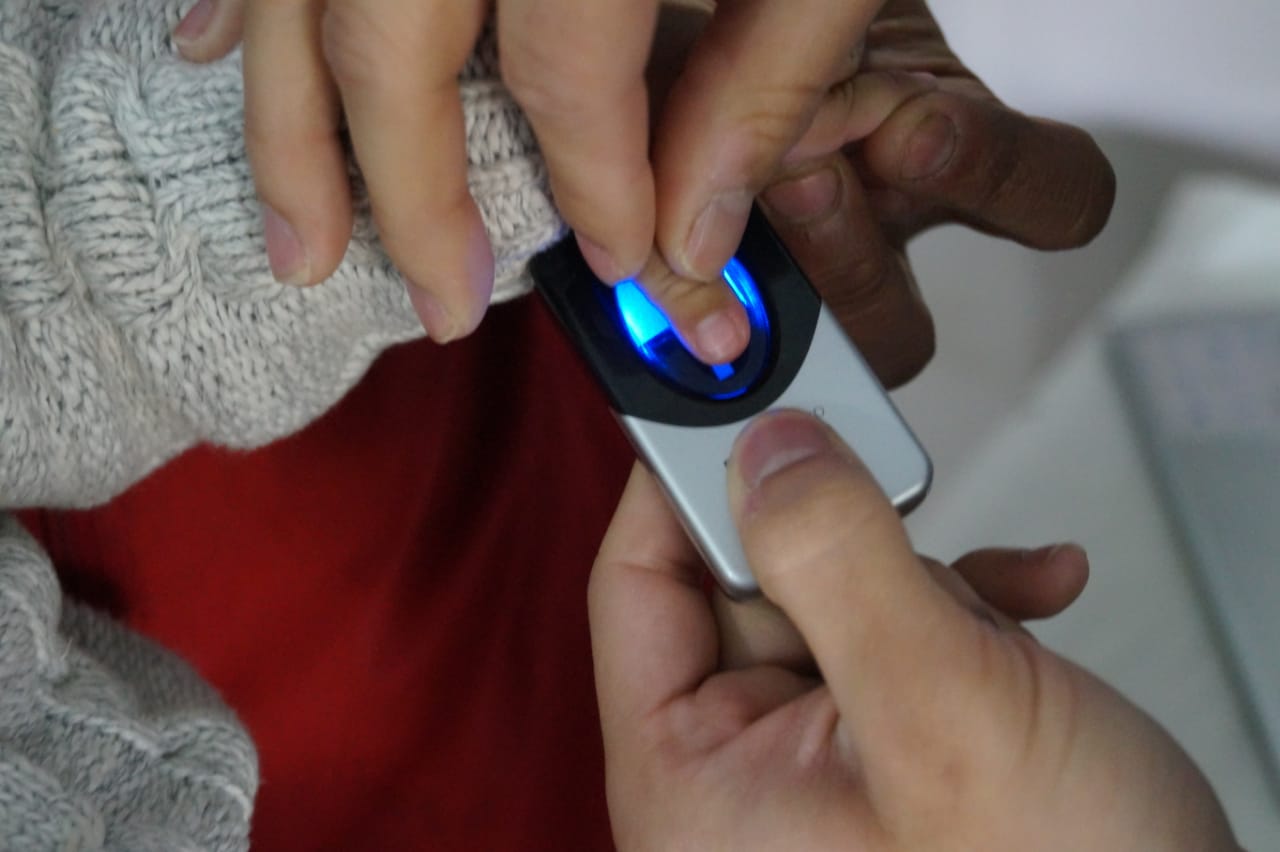}}\hfill
    \subfloat[1,900 ppi RaspiReader]{\includegraphics[height=1.21in]{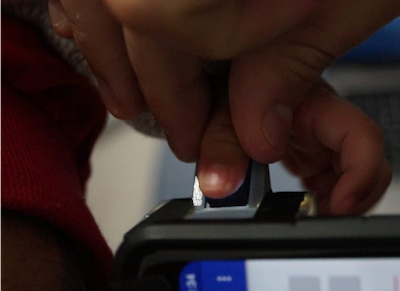}}\\
     \subfloat[]{\includegraphics[width=0.45\linewidth]{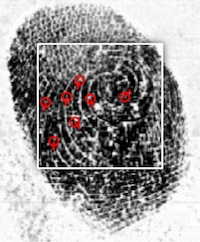}}\hfill
    \subfloat[]{\includegraphics[width=0.45\linewidth]{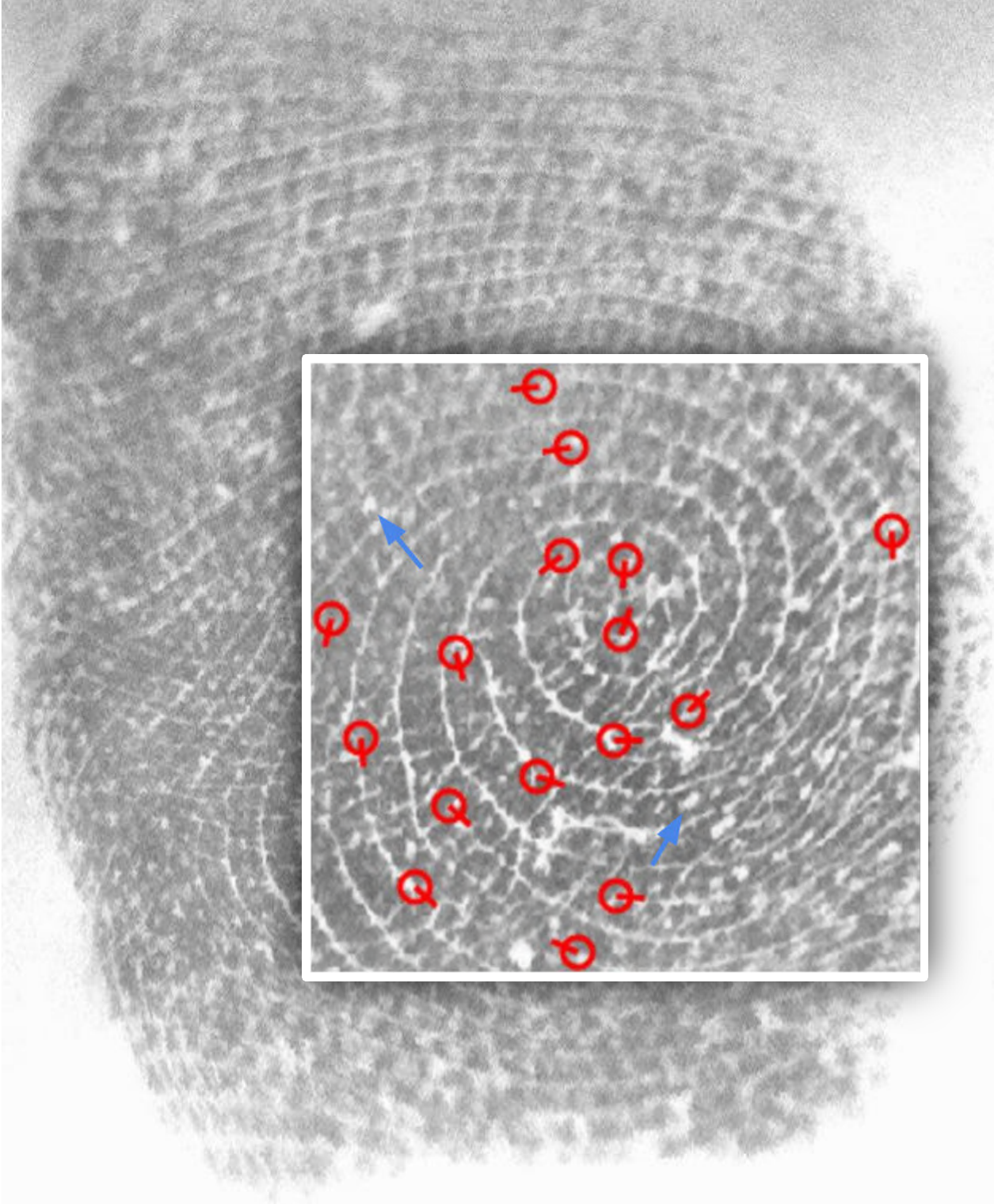}}
    \caption{An infant's fingerprints are acquired via (a) a 500 ppi commercial reader (Digital Persona U.are.U 4500)  and (b) our custom 1,900 RaspiReader. The captured fingerprint images of the right thumb from the commercial reader and the Infant-Prints reader for a 13 day old infant are shown in (c) and (d), respectively. Manually annotated minutiae are shown in red circles (location) with a tail (orientation). Blue arrows denote pores on the ridges.}
    \label{fig:infant_fpt}
\end{figure}

\section{High-Resolution Fingerprint Reader}
Almost all the fingerprint readers used in government and commercial applications capture images at a resolution of 500 ppi. This resolution is sufficient to resolve adult fingerprint ridges that have an inter-ridge  spacing of about 8-10 pixels. However, 500 ppi resolution is not adequate for infant fingerprint capture since infant fingerprints have an inter-ridge spacing of 4-5 pixels (sometimes the width of a valley may be less than 1 pixel for an infant fingerprint captured at 500 ppi). 

Some cheaper readers (50 USD) reach 1,000 ppi only after upsampling  the fingerprint image~\cite{silk}. However, Jain \textit{et al.}~\cite{jain} showed that even at a native resolution\footnote{Native resolution is the resolution at which the sensor is capable of capturing (no upsampling or downsampling).} of 1,270 ppi, fingerprint recognition of young infants (0-6 months) was much lower than infants 6 months and older. The lack of an affordable, compact and high resolution fingerprint reader motivated us to construct a first-of-a-kind, 1,900 ppi fingerprint reader, called RaspiReader~(Fig.~\ref{fig:reader}), enabling capture of high-fidelity infant fingerprints (Fig.~\ref{fig:infant_fpt}), particularly those in the age range 0-3 months. Unlike our prior efforts to build a compact and cheap reader for adults~\cite{engelsma1,engelsma2}, both the cost and size of the infant fingerprint reader has been significantly reduced  (from 180 USD to 85 USD and $4"\times4"\times4"$ to $1"\times2"\times3"$). Furthermore, the fingerprint reader is now more ergonomic for infant fingerprints since it has a glass prism towards the front of the reader (Fig.~\ref{fig:reader}) rather than flush with the top of the reader (as is the case with commercial readers). Since infants frequently clench their fists and have very short fingers, placing the prism out front significantly eases placement of an infant's finger on the platen (Fig.~\ref{fig:infant_fpt} (b)).

The entire design and 3D parts for the reader casing along with step by step assembly instructions are open sourced.\footnote{\url{https://github.com/engelsjo/RaspiReader}} Figure~\ref{fig:infant_fpt} shows that this custom 1,900 ppi fingerprint reader is able to capture (500 millisecond capture time) the minute friction ridge pattern of a 13 day old infant (both minutiae and pores) with higher fidelity than the 500 ppi Digital Persona U.are.U. 4500 reader.

\begin{figure}[!t]
    \centering
    \subfloat[]{\includegraphics[width=0.8\linewidth]{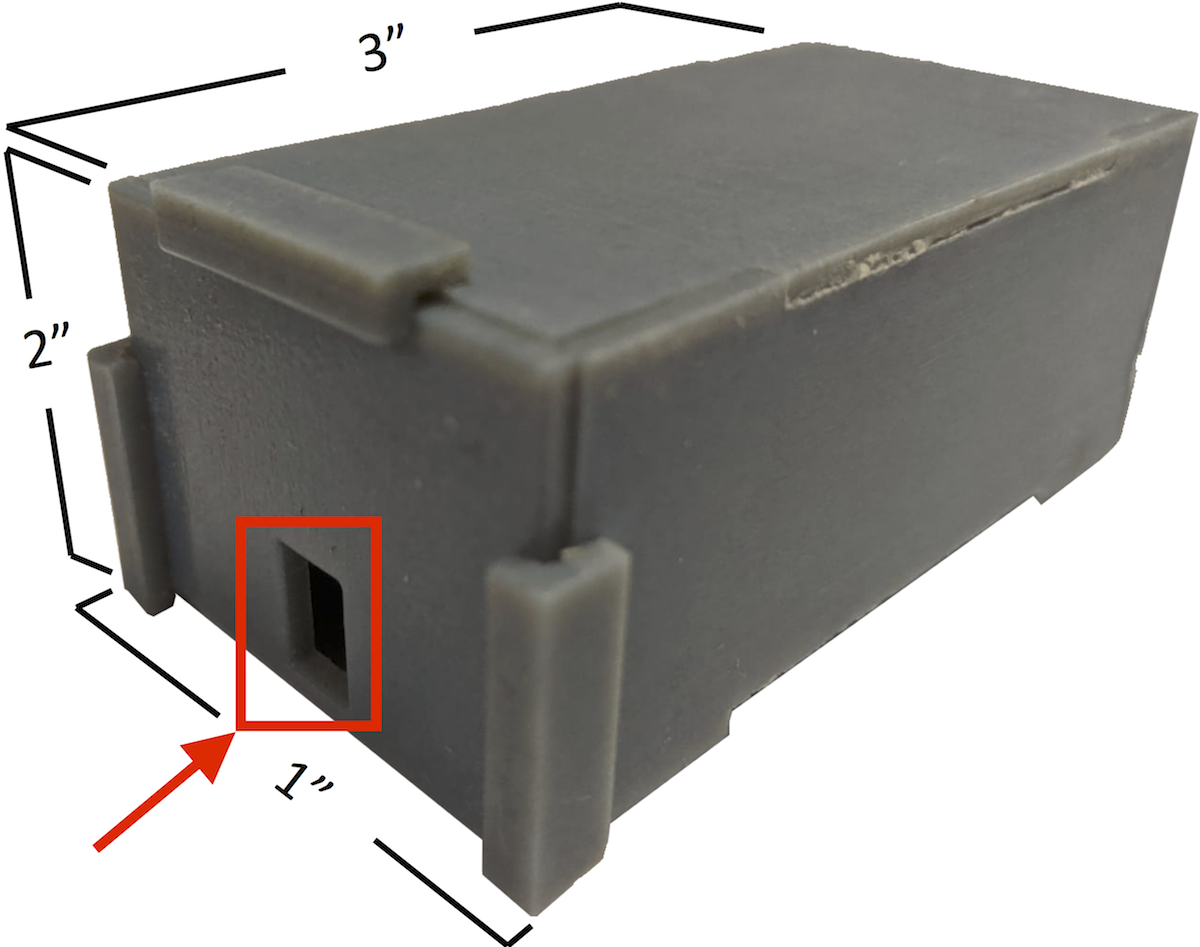}}\\
    \subfloat[Contactless image]{\includegraphics[width=0.45\linewidth]{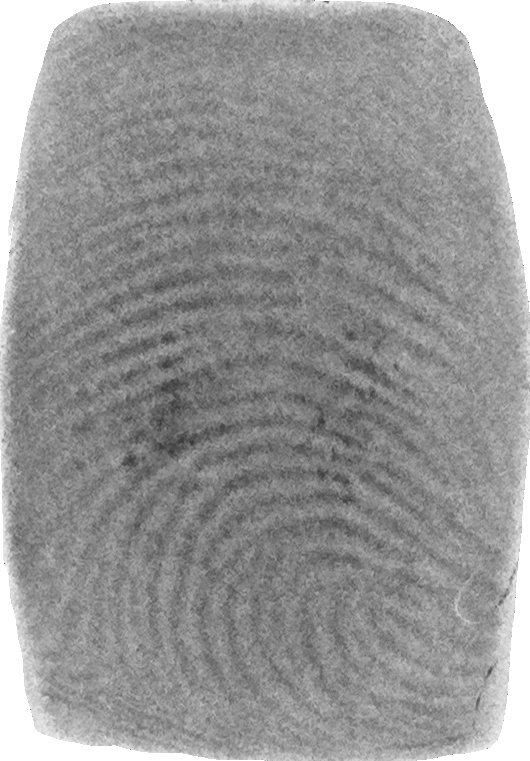}}\hfill
    \subfloat[Contact-based image]{\includegraphics[width=0.45\linewidth]{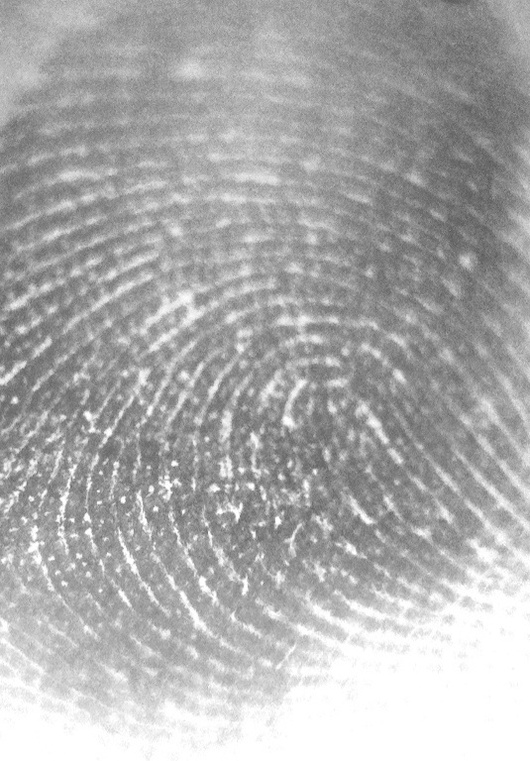}}
    \caption{(a) Prototype of our 1,900 ppi contactless fingerprint reader. During capture, an infant's finger is placed on top of a small, contactless, rectangular opening (annotated in red) on the reader (the size of this opening can be adjusted with different sized slots). A camera captures the infant's fingerprint from behind the rectangular opening. Examples of a processed (segmented, contrast enhanced), contactless infant thumb-print (2 months old) is shown in (b) and the same infant's thumb-print acquired via contact-based fingerprint reader in (c).}
    \label{fig:contactless_reader}
\end{figure}

We also prototype a contactless variant of our contact-based infant fingerprint reader. Similar to~\cite{gates}, we adopt a different size finger rest for different size thumbs. In this manner, we are able to compare contact-based high resolution fingerprint readers with the high resolution contactless sensing technology. Figure~\ref{fig:contactless_reader} shows an example infant fingerprint captured by both our contactless and contact-based fingerprint reader. 

\section{Longitudinal Fingerprint Dataset}
To effectively demonstrate the utility of an infant fingerprint recognition system for the applications we have highlighted above, we must be able to show its ability to recognize a child based on fingerprints acquired months after the initial enrollment. Such an evaluation requires a longitudinal fingerprint dataset which contains fingerprint images of the same infant over time at successive intervals. Collecting such a dataset is a significant challenge as it requires the cooperation of an infant's parents in returning to the clinic multiple times for participation in the study. It also requires working with uncooperative infants who may become hungry or agitated during the data collection (our ergonomic fingerprint reader alleviated some of these challenges). 

\begin{figure}[!t]
    \centering
    \includegraphics[width=\linewidth]{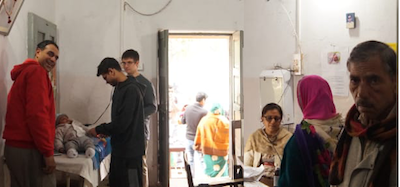}
    \caption{Infant fingerprint collection at Saran Ashram hospital, Dayalbagh, India. Pediatrician, Dr. Anjoo Bhatnagar, explaining longitudinal fingerprint study to the mothers while the authors are acquiring an infant's fingerprints in her clinic. Parents also sign a consent form approved by the Institutional Review Board (IRB) of our organizations.}
    \label{fig:data_collection}
\end{figure}

We have collected a dataset comprised of longitudinal fingerprint images of 315 infants (all enrolled at 0-3 months of age) at the Saran Ashram hospital in Dayalbagh, India across four sessions (see Fig.~\ref{fig:data_collection})\footnote{Our dataset collection was approved by the Institutional Review Board (IRB) of Michigan State University and ethics committee of Dayalbagh Educational Institute and Saran Ashram Hospital. The fingerprint dataset cannot be made publicly available per the IRB regulations and parental consents.}:
\begin{enumerate}
    \item Session 1: December 12-19, 2018
    \item Session 2: March 3-9, 2019
    \item Session 3: September 12-21, 2019
    \item Session 4: January 17-24, 2020
\end{enumerate}
The infants were patients of the pediatrician, Dr. Anjoo Bhatnagar (Fig.~\ref{fig:data_collection}). Prior to data collection, the parents were required to sign a consent form (approved by authors' institutional review board and the ethics committee of Saran Ashram hospital).

In a single session, we attempted to acquire a total of two impressions per thumb (sometimes we captured more (\textit{e.g.} 4 impressions) or less (\textit{e.g.} 1 impression) depending on the cooperative nature of the infant). Although incentive was offered to parents for their data collection efforts, it was often difficult for them to meet our fingerprint capture schedule because of festivals, vacations, moving to a different city or loss of interest in the project. For this reason, out of the 315 total infants that we encountered, 25 infants were present in all four sessions, 54 infants came to only three sessions, 109 infants came to only two sessions, and 127 infants came to only one session. During collection, a dry or wet wipe was used, as needed, to clean the infant's finger prior to fingerprint acquisition. On average, data capture time, for 4 fingerprint images (2 per thumb) and a face image per infant, was 3 minutes\footnote{Data capture time includes parents signing the consent forms, record-keeping, and pacifying non-cooperative infants.}. This enabled a reasonably high throughput during the in-situ evaluation, akin to the operational scenario in immunization and nutrition distribution centers. Longitudinal fingerprint dataset statistics are given in Table~\ref{tab:dataset_statistics}.

\newcolumntype{Y}{>{\centering\arraybackslash}X}
\begin{table}[!t]
\centering
\caption{Infant Longitudinal Fingerprint Dataset Statistics}
\label{tab:dataset_statistics}
\begin{threeparttable}
\begin{tabularx}{\linewidth}{l|Y}
\noalign{\hrule height 1.1pt}
\textbf{\# Sessions} & 4\\ \hline
\textbf{\# Infants}  & 315\\ \hline
\textbf{Total \# images} & 3,071 \\ \hline
\textbf{Age at enrollment} & 0 - 3 mos.\\  \hline
\textbf{\# Subjects with no time lapse\tnote{*}} & 127\\ \hline
\textbf{\# Subjects with 3 months lapse\tnote{*}} & 121\\ \hline
\textbf{\# Subjects with 6 months lapse\tnote{*}} & 29\\ \hline
\textbf{\# Subjects with 9 months lapse\tnote{*}} & 101\\ \hline
\textbf{\# Subjects with 12 months lapse\tnote{*}} & 41\\ \hline
\textbf{Male to Female Ratio} & 43\% to 57\%\\
\noalign{\hrule height 1.1pt}
\end{tabularx}
\begin{tablenotes}\footnotesize
\item[*] Time lapse between enrollment and authentication image.
\end{tablenotes}
\end{threeparttable}
\end{table}

\section{Infant Fingerprint Matching}

State-of-the-art fingerprint feature extractors and matchers are designed to operate on 500 ppi adult fingerprint images. This limitation forced the authors in~\cite{jain} to down-sample the fingerprint images captured at 1,270 ppi to enable compatibility with COTS (Commercial Off The Shelf) matchers. The authors in~\cite{gates} also had to down-sample images captured at 3,400 ppi in order to make them compatible with adult fingerprint matching systems. In our preliminary study~\cite{infant_prints}, we developed a custom Convolutional Neural Network (CNN) based texture-matcher which directly operates on 1,900 ppi fingerprint images so that we did not have to down-sample images and discard valuable discriminative cues available in high resolution images. The final matching score in~\cite{infant_prints} was based on the fusion of (i) our CNN-based custom texture matcher and (ii) two state-of-the-art COTS matchers.

In this work, we (i) incorporate an enhancement and fingerprint aging preprocessing module, (ii) improve our high-resolution texture matcher from~\cite{infant_prints}, and (iii) propose a high-resolution minutiae extractor trained on manually annotated infant fingerprint images. Combining these algorithmic improvements with two state-of-the-art fingerprint matchers (a latent fingerprint matcher, and a minutiae matcher) enables us to improve our recognition accuracy over that which was reported in our preliminary study~\cite{infant_prints}. In the following subsections, we discuss in more detail each of these algorithmic improvements.

\subsection{Minutiae Matcher}

Our high resolution minutiae matcher is comprised of (i) a high-resolution minutiae extractor, (ii) a minutiae aging model, and (iii) the Verifinger v10.0 ISO minutiae matcher. In the following subsections, we describe each of these algorithmic components. 

\subsection{Minutiae Extraction}

Recent approaches to minutiae extraction in the literature have found that deep networks are capable of delivering superior minutiae extraction performance in comparison to traditional approaches~\cite{minutiae1, minutiae2, minutiae3, minutiae4}. Furthermore, the authors in~\cite{Cao1} showed that deep learning based minutiae extractors are particularly well suited for low quality fingerprint images such as latent fingerprint images. Since infant fingerprints can also be regarded as a ``low-quality" fingerprint (heavy non-linear distortion, motion blur from uncooperative subjects, small inter-ridge spacing, very moist or dry fingers, dirty fingers), we choose to adopt the deep learning based minutiae extraction approach from~\cite{Cao1} (with modifications) for high-resolution infant minutiae extraction. In our experiments, we demonstrate that the high-resolution minutiae extractor is capable of boosting the infant fingerprint recognition performance. 

The core of the minutiae extraction algorithm proposed in~\cite{Cao1} is a fully-convolutional auto-encoder $M(.)$ which is trained to regress from an input fingerprint image $\mathbf{I} \in \mathbb{R}^{n\times m}$ to a ground truth minutiae map $\mathbf{H} \in \mathbb{R}^{n\times m \times 12}$ via $\mathbf{\hat{H}} = M(I)$, where $\mathbf{\hat{H}}$ is the predicted minutiae map. The spatial locations of hot spots in the minutiae map indicate the locations of minutiae points, and the 12 different channels of the minutiae map encode the orientation of the minutiae points. The parameters of $M$ are trained in accordance with Equation~(\ref{eq:mmap}).

\begin{equation}
\label{eq:mmap}
    \mathcal{L}_{minutiae} = ||\mathbf{\hat{H}} - \mathbf{H}||^2_2
\end{equation}

This estimated 12 channel minutiae map $\mathbf{\hat{H}}$ can be subsequently converted into a variable length minutiae set $\{(x_1, y_1, \theta_1), ..., (x_N, y_N, \theta_N)\}$ with $N$ minutiae points via an algorithm which locates local maximums in the channels (locations) and individual channel contributions (orientations) followed by non-maximal suppression to remove spurious minutiae~\cite{Cao1}.

To obtain ground truth minutiae maps $\mathbf{H}$ for computing $\mathcal{L}_{minutiae}$, we encode a ground truth minutiae set for a given infant fingerprint following the approach of~\cite{Cao1} for latent fingerprints. In particular, given a ground truth minutiae set $T = \{m_1, m_2, ..., m_N \}$ with $N$ minutiae and $m_t = (x_t, y_t, \theta_t)$, $\mathbf{H}$ at position $(i, j, k)$ is given by:

\begin{equation}
\label{eq:main_mmap}
H(i, j, k) = \sum_{t=1}^N C_s((x_t,y_t), (i, j)) \cdot C_o(\theta_t, 2k\pi /12)
\end{equation} where $C_s(.)$ is the spatial contribution and $C_o(.)$ is the orientation contribution of minutiae point $m_t$ to the minutiae map at $(i, j, k)$. Note, $C_s(.)$ is based upon the euclidean distance of $(x_t,y_t)$ to $(i,j)$ and $C_o(.)$ is based on the orientation difference between $\theta_t$ and $2k\pi /12$ as follows:

\begin{equation}
\label{eq:contribs_dist}
C_s((x_t,y_t), (i,j)) = exp(-\frac{||(x_t,y_t)-(i,j)||_2^2}{2\sigma_s^2})
\end{equation}

\begin{equation}
\label{eq:contribs_theta}
C_o(\theta_t,2k\pi /6) = exp(-\frac{d\phi(\theta_t,2k\pi /12)}{2\sigma_s^2})
\end{equation} where $\sigma_s^2$ is a parameter controlling the width of the gaussian, and $d\phi(\theta_1,\theta_2)$ is difference in orientation between angles $\theta_1$ and $\theta_2$:

\begin{equation}
\label{eq:theta_diffs}
d\phi(\theta_1, \theta_2) = \begin{cases}
|\theta_1 - \theta_2| & -\pi \leq \theta_1 - \theta_2 \leq \pi \\
2\pi - |\theta_1 - \theta_2| & otherwise.
\end{cases}
\end{equation}

\begin{figure*}[!t]
    \centering
    \includegraphics[scale=0.725]{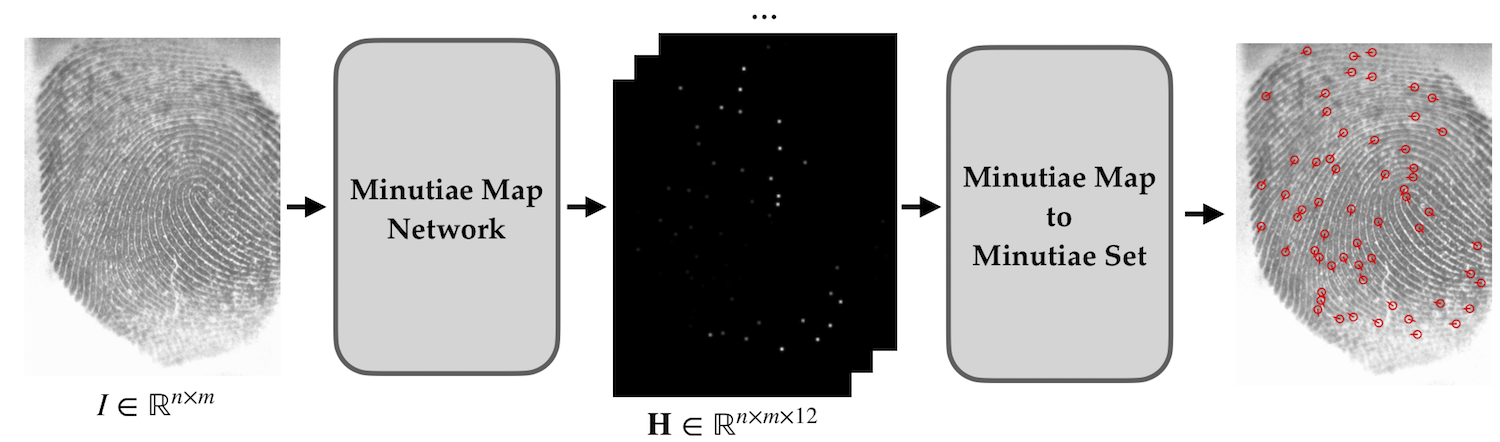}
    \caption{Overview of the minutiae extraction algorithm. An input fingerprint of any size ($n\times m$) is passed to the minutiae extraction network (Table~\ref{table:minutiae_network}). The network outputs a $n\times m\times12$ minutiae map $\mathbf{H}$ which encodes the minutiae locations and orientations of the input fingerprint. Finally, the minutiae map is converted to a minutiae set $\{(x_1, y_1, \theta_1), ..., (x_N, y_N, \theta_N) \}$ of $N$ minutiae.}
    \label{fig:minutiae_extractor}
\end{figure*}

\begin{figure}[!t]
    \centering
    \subfloat[]{\includegraphics[width=0.49\linewidth]{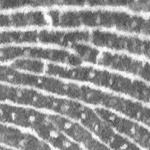}}\hfill
    \subfloat[]{\includegraphics[width=0.49\linewidth]{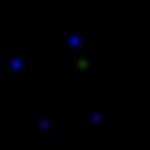}}\\
    \caption{An example infant fingerprint patch (a) and the corresponding minutiae map (b). Note, we only show 3 channels of the 12 channel minutiae map here for illustrative purposes (red channel is the first channel, green is the fifth channel, and blue is the ninth channel). Given the full 12 channels of the minutiae map in (b), we can compute the minutiae locations $(x, y)$ and orientations $\theta$ of the 1,900 ppi fingerprint patch in (a).}
    \label{fig:mmap_eg}
\end{figure}

\newcommand{\specialcell}[2][c]{%
  \begin{tabular}[#1]{@{}c@{}}#2\end{tabular}}
  \newcommand{\tabitem}{~~\llap{\textbullet}~~}

\begin{table}[t]
\caption{Minutiae Extraction Network}
 \centering
\begin{threeparttable}
\begin{tabular}{c c c}
 \toprule
 Type & \specialcell{Output Size} & \specialcell{Filter Size, Stride} \\
 \toprule
 Convolution & $256\times256\times64$ & $4\times4$, $1$\\
 \midrule
 Convolution & $128\times128\times64$ & $4\times4$, $2$\\
  \midrule
 Convolution & $64\times64\times128$ & $4\times4$, $2$ \\
  \midrule
 Convolution & $32\times32\times256$ & $4\times4$, $2$\\
  \midrule
 Convolution & $16\times16\times384$ & $4\times4$, $2$ \\
  \midrule
 Convolution & $8\times8\times512$ & $4\times4$, $2$ \\
  \midrule
 Convolution & $8\times8\times1024$ & $4\times4$, $1$ \\
  \midrule
 Convolution & $4\times4\times1024$ & $4\times4$, $2$ \\
  \midrule
 Deconvolution & $4\times4\times1024$ & $4\times4$, $1$ \\
   \midrule
 Deconvolution & $8\times8\times512$ & $4\times4$, $2$ \\
   \midrule
 Deconvolution & $16\times16\times384$ & $4\times4$, $2$ \\
   \midrule
 Deconvolution & $32\times32\times256$ & $4\times4$, $2$ \\
   \midrule
 Deconvolution & $64\times64\times128$ & $4\times4$, $2$ \\
   \midrule
 Deconvolution & $128\times128\times64$ & $4\times4$, $2$ \\
   \midrule
 Deconvolution & $256\times256\times32$ & $4\times4$, $2$ \\
   \midrule
 Deconvolution & $256\times256\times12$ & $4\times4$, $1$ \\
 \bottomrule
\end{tabular}
\begin{tablenotes}
\item[\textdagger] During training, input patches are $256\times256$. During testing, the input can be of any size (the network is fully convolutional).
\end{tablenotes}
\end{threeparttable}
\label{table:minutiae_network}
\end{table}

An example infant fingerprint patch, and a few channels of its 12 channel ground truth minutiae map are shown in Figure~\ref{fig:mmap_eg}. An overview of our end-to-end minutiae extraction algorithm is shown in Figure~\ref{fig:minutiae_extractor}. In contrast to the 500 ppi latent fingerprint minutiae extractor in~\cite{Cao1}, we directly train our minutiae extractor on infant fingerprint patches at 1,900 ppi resolution. In this manner, we do not remove any discriminative cues (via down-sampling) from the input infant fingerprint images prior to performing minutiae extraction. Operating at a high resolution requires a deeper network architecture than that which was utilized in~\cite{Cao1}. Our network architecture is shown in detail in Table~\ref{table:minutiae_network}. Note that while we train our auto-encoder on infant fingerprint \textit{patches}, during test time, we  input a full size infant fingerprint (of varying width and height) since our architecture is \textit{fully-convolutional} and as such, is amenable to different size inputs.

\subsubsection{Manual Minutiae Markup for Training}

\begin{figure}[!h]
    \centering
    \includegraphics[scale=1.0]{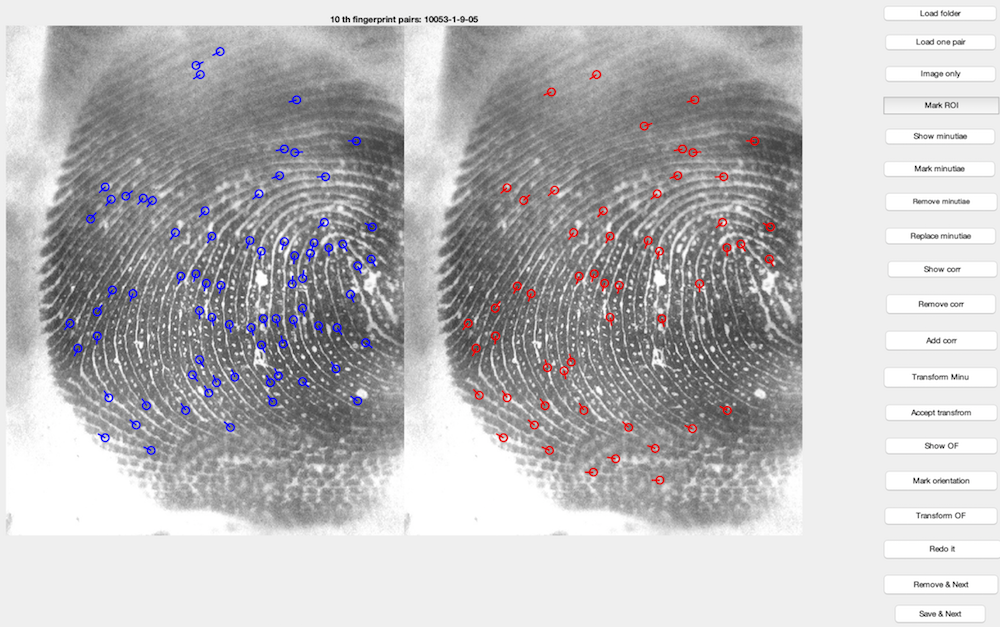}
    \caption{View of the manual minutiae markup/editing software used to markup minutiae locations on a subset of fingerprint images. These markups were later used as ground truth to train our high resolution infant minutiae extractor. The fingerprint on the left (blue annotations) is coarsely annotated with Verifinger v10 SDK to help speed up the annotation process. The fingerprint on the right (red annotations) are the manually edited minutiae.}
    \label{fig:minutiae_markup}
\end{figure}

As seen in the previous section from Equations~\ref{eq:main_mmap}-\ref{eq:theta_diffs}, obtaining ground truth minutiae maps $\mathbf{H}$ for training our minutiae map extraction network $M(.)$ requires a ground truth minutiae set $T$ for each input infant fingerprint. To obtain these ground truth minutiae sets for training, we manually annotate the minutiae locations and orientations of $610$ infant fingerprints in our dataset for which we had limited longitudinal data (\textit{i.e.} the infant only visited 1 or 2 sessions). These fingerprints are separated from our evaluation dataset. We manually annotated the infant fingerprints using the GUI tool shown in Figure~\ref{fig:minutiae_markup}. The tool enables the addition of new minutiae and the removal of spurious minutiae. To make the markup task easier, we first automatically annotate the minutiae points on the $610$ infant fingerprints using the Verifinger v10.0 minutiae extraction SDK. Then, we manually refine the Verifinger annotations with our markup GUI. Each manually annotated fingerprint was reviewed multiple times by one of 4 experts in the field of fingerprint recognition.

\begin{figure*}[!h]
    \centering
    \subfloat[]{\includegraphics[width=0.24\linewidth]{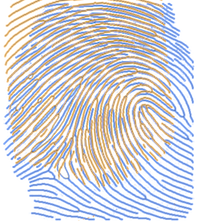}}\hfill
    \subfloat[]{\includegraphics[width=0.24\linewidth]{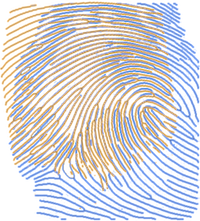}} \hfill
    \subfloat[]{\includegraphics[width=0.24\linewidth]{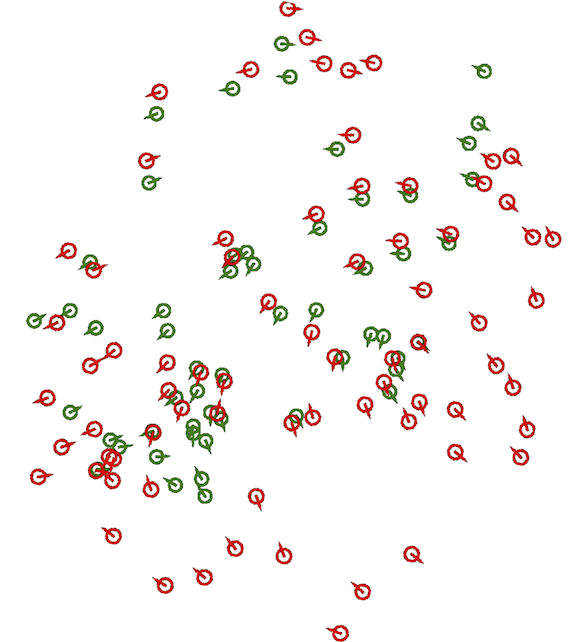}} \hfill
    \subfloat[]{\includegraphics[width=0.24\linewidth]{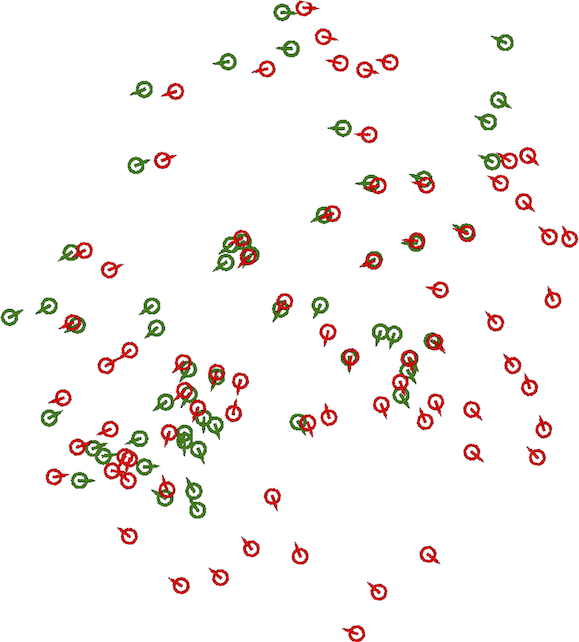}}
    \caption{Effects of aging. (a) Acquired 3 month old enrollment image (orange) is overlaid on a 1 year old probe image (blue). (b) An aged 3 month old enrollment image (orange) is overlaid on a 1 year old probe image (blue). (c) 3 month old enrollment minutiae set (green) is overlaid on a 1 year old probe minutiae set (red). (d) An aged 3 month old enrollment minutiae set (green) is overlaid on a 1 year old probe minutiae set (red). Following aging (b, d), the enrollment image and probe image (and corresponding minutiae sets) overlap better.}
    \label{fig:aging}
\end{figure*}

While the $610$ manually annotated infant fingerprints provide an accurate ground truth dataset for training our minutiae extraction network, in practice, it is insufficiently small for training a deep network (Table~\ref{table:minutiae_network}). Therefore, rather than training our minutiae extraction network from scratch on the $610$ manually annotated infant fingerprints, we first pretrain our minutiae extraction network on $9,508$ infant/child fingerprints collected in~\cite{infant_jain} and coarsely annotated with minutiae using the Verifinger v10.0 minutiae extractor. After pretraining our minutiae extraction network on these $9,508$ coarsely annotated (using Verifinger) fingerprints, we finally fine-tune all parameters of our network (Table~\ref{table:minutiae_network}) using our more accurate $610$ manually annotated ground truth infant fingerprint images ($560$ used for training, $50$ used for validation). We optimize our network parameters using the Adam optimizer and weight decay set to $4\times10^{-5}$. When training the network on the $9,508$ coarsely annotated training data, we use a learning rate of $0.01$. When fine-tuning our network (all parameters fine-tuned) on our manually annotated fingerprint images, we reduce the learning rate to $0.0001$. We use the minutiae detection accuracy on our 50 manually annotated validation fingerprints as a stopping criteria for the training. Finally, our network is trained on $256\times256$ patches to increase the number of training samples, and we employ data augmentations such as random rotations, cropping, translations, and flipping. 

\begin{figure*}[!t]
    \centering
    \subfloat[]{\includegraphics[width=0.19\linewidth]{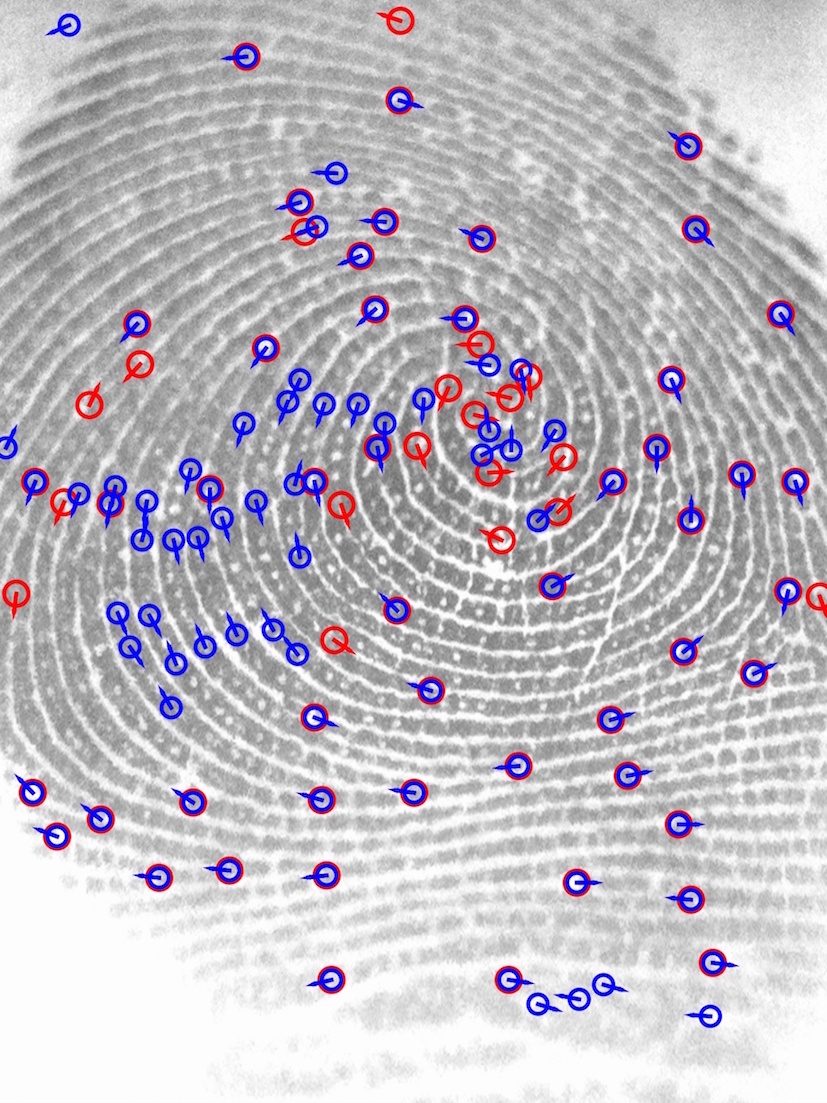}}\hfill
    \subfloat[]{\includegraphics[width=0.19\linewidth]{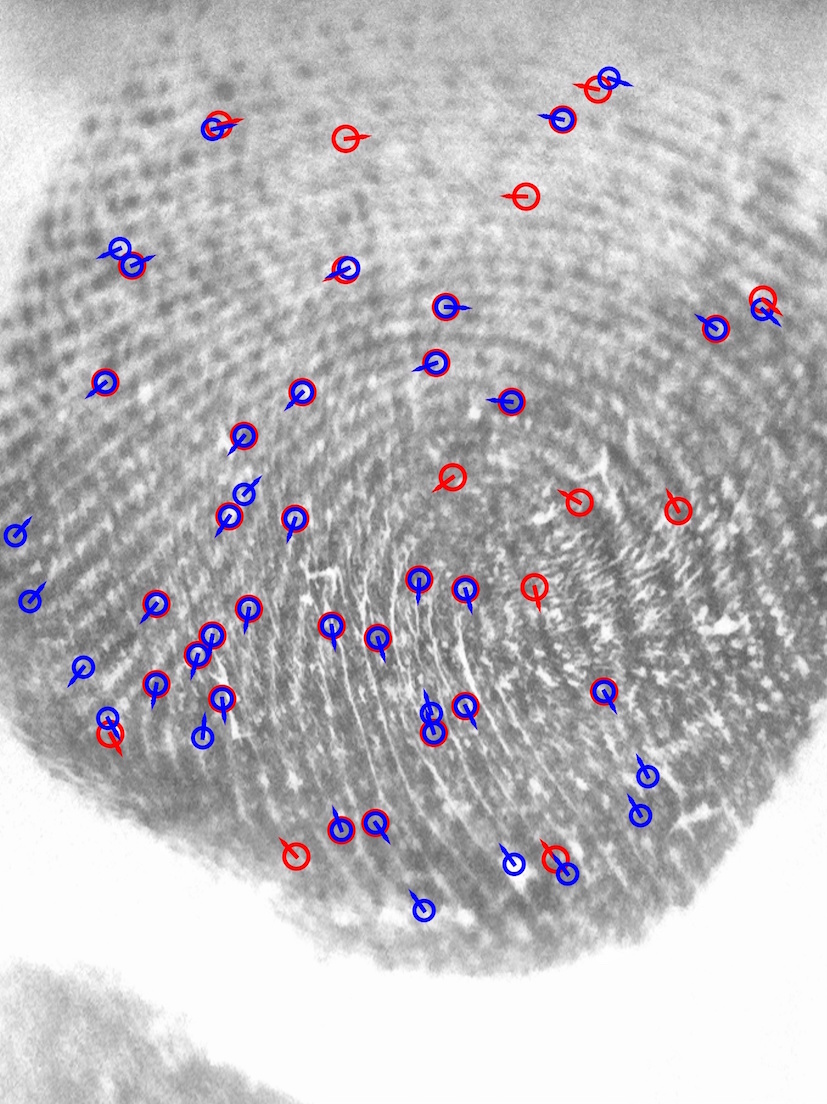}}\hfill
    \subfloat[]{\includegraphics[width=0.19\linewidth]{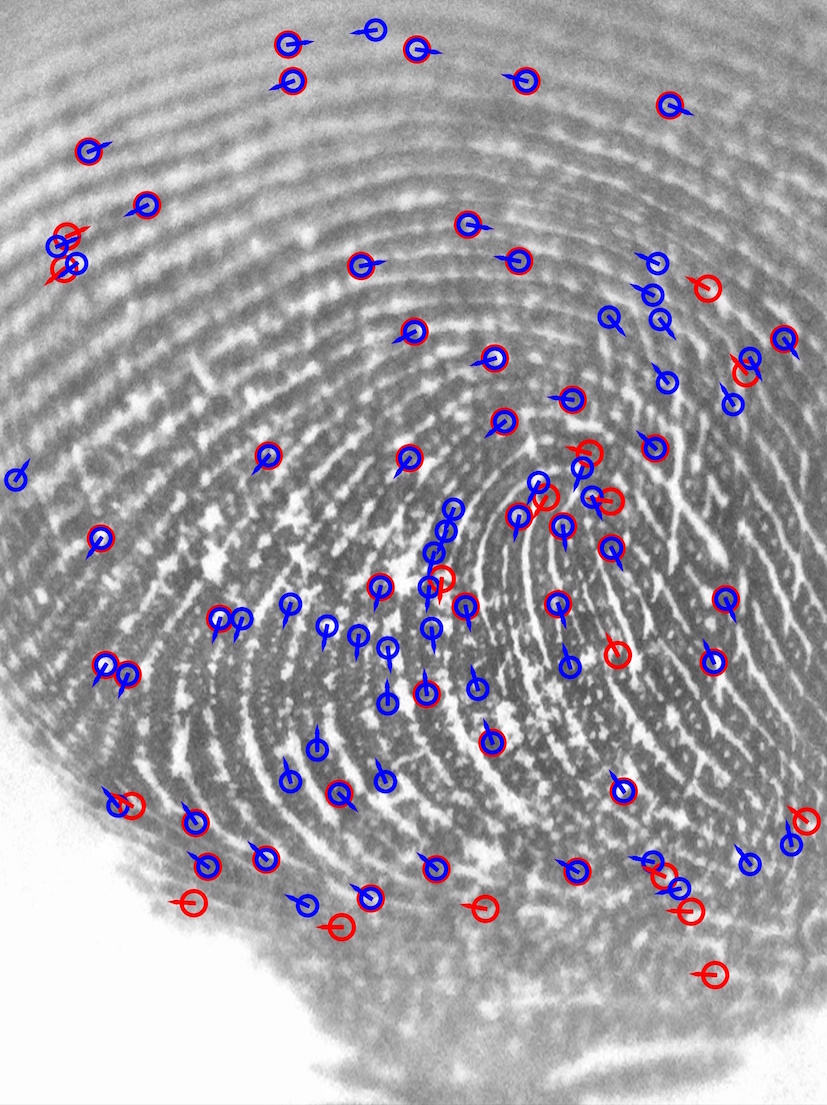}}\hfill
    \subfloat[]{\includegraphics[width=0.19\linewidth]{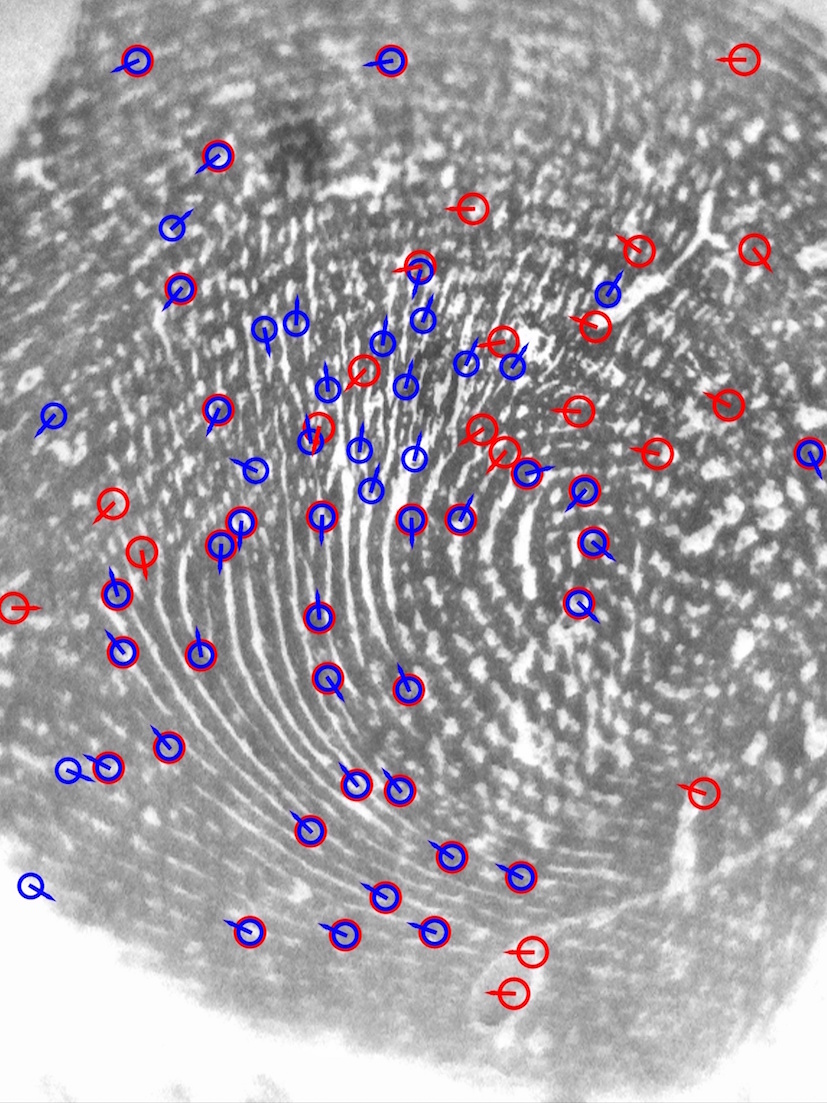}}\hfill
    \subfloat[]{\includegraphics[width=0.19\linewidth]{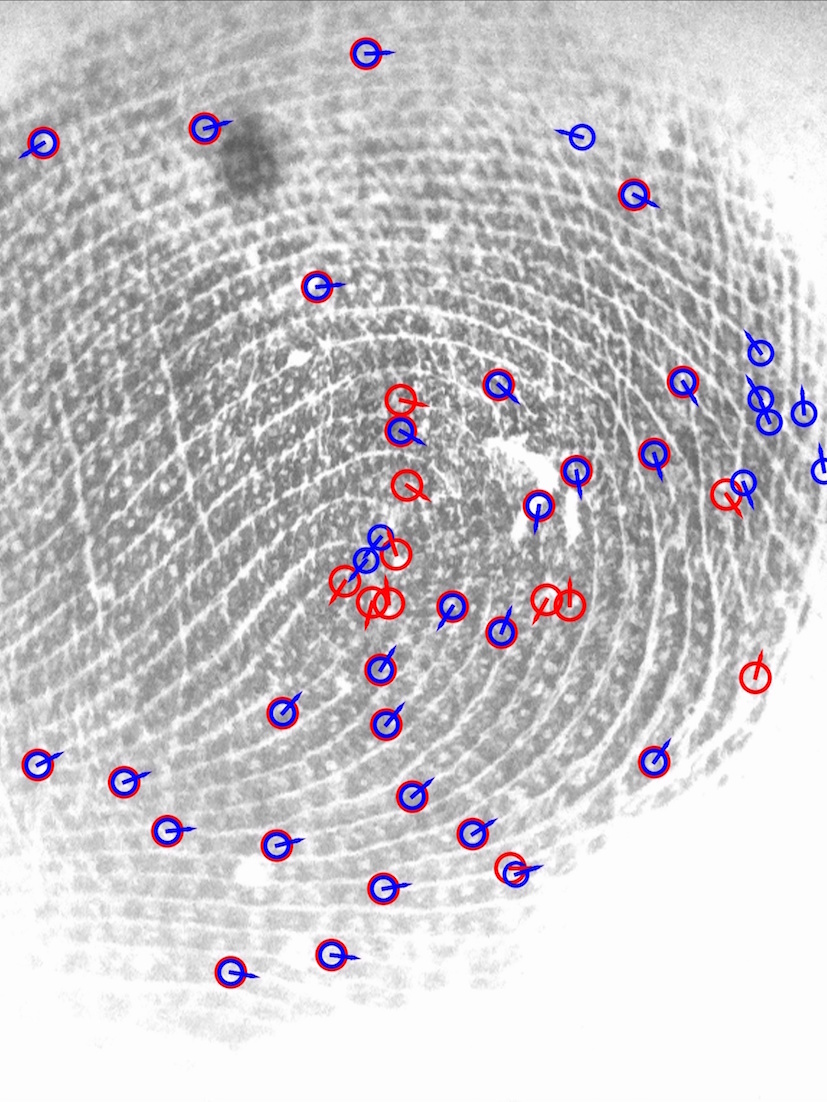}}\\
    \subfloat[]{\includegraphics[width=0.19\linewidth]{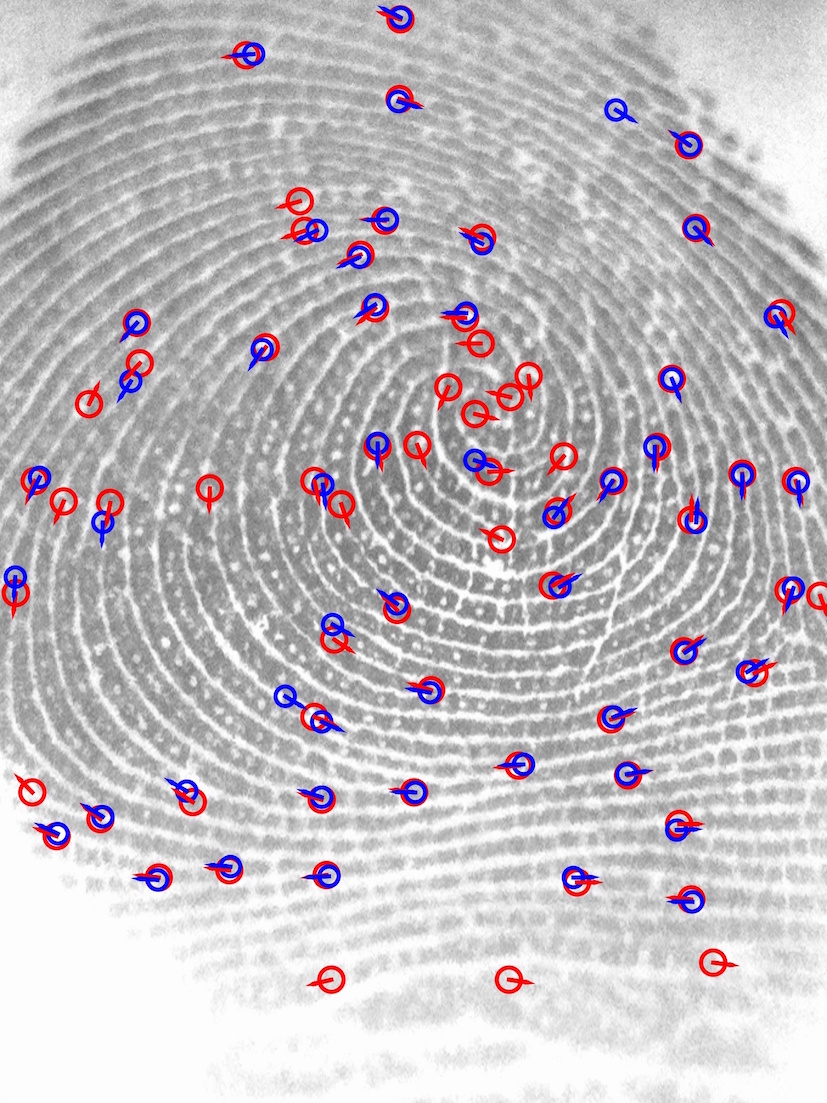}}\hfill
    \subfloat[]{\includegraphics[width=0.19\linewidth]{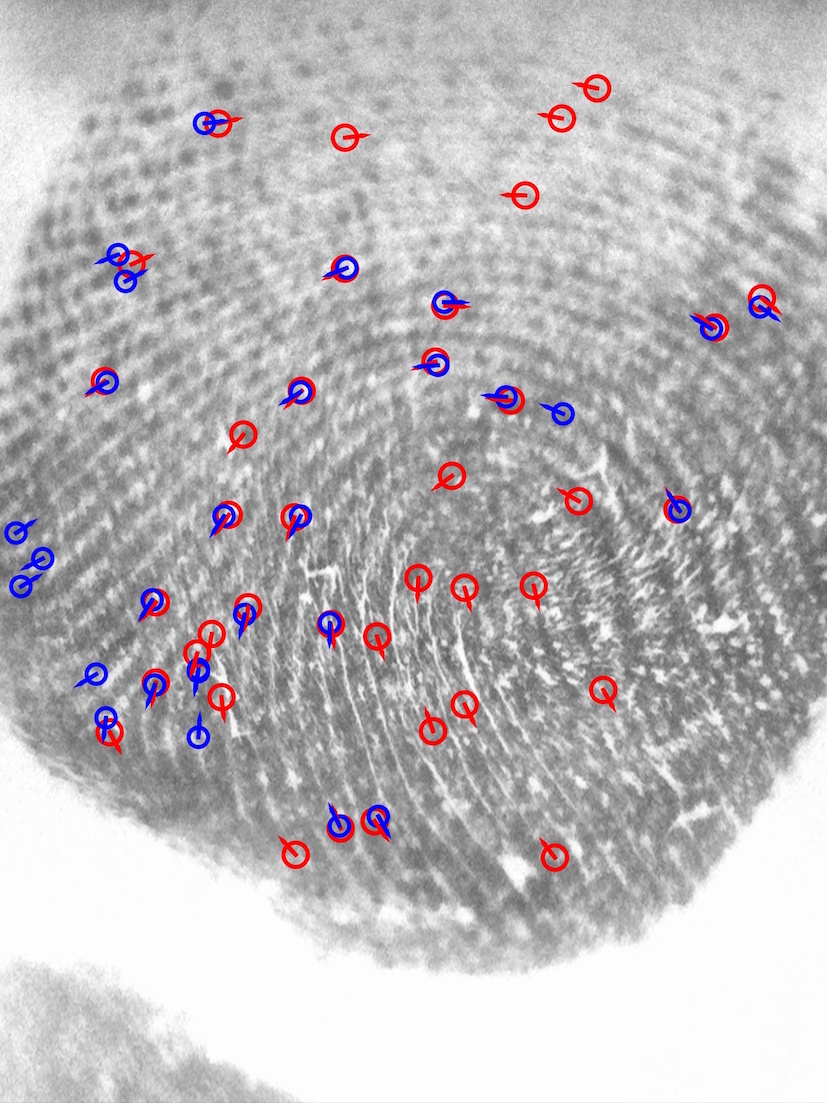}}\hfill
    \subfloat[]{\includegraphics[width=0.19\linewidth]{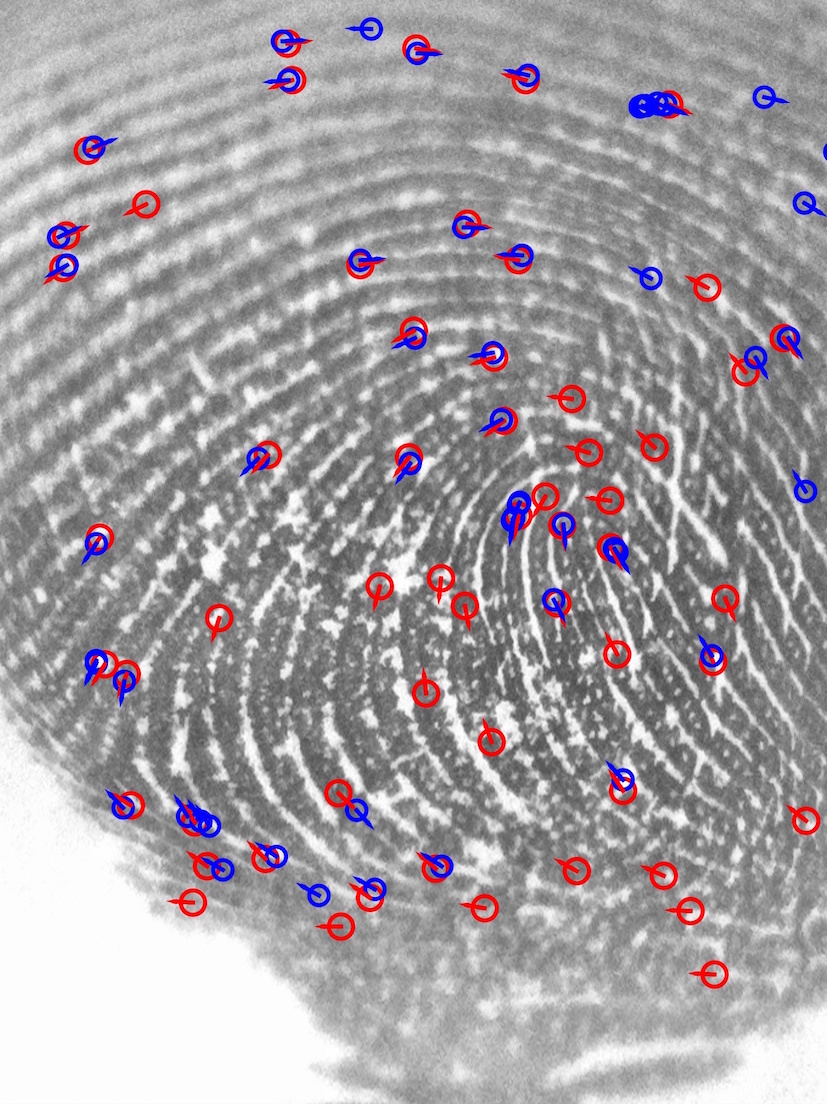}}\hfill
    \subfloat[]{\includegraphics[width=0.19\linewidth]{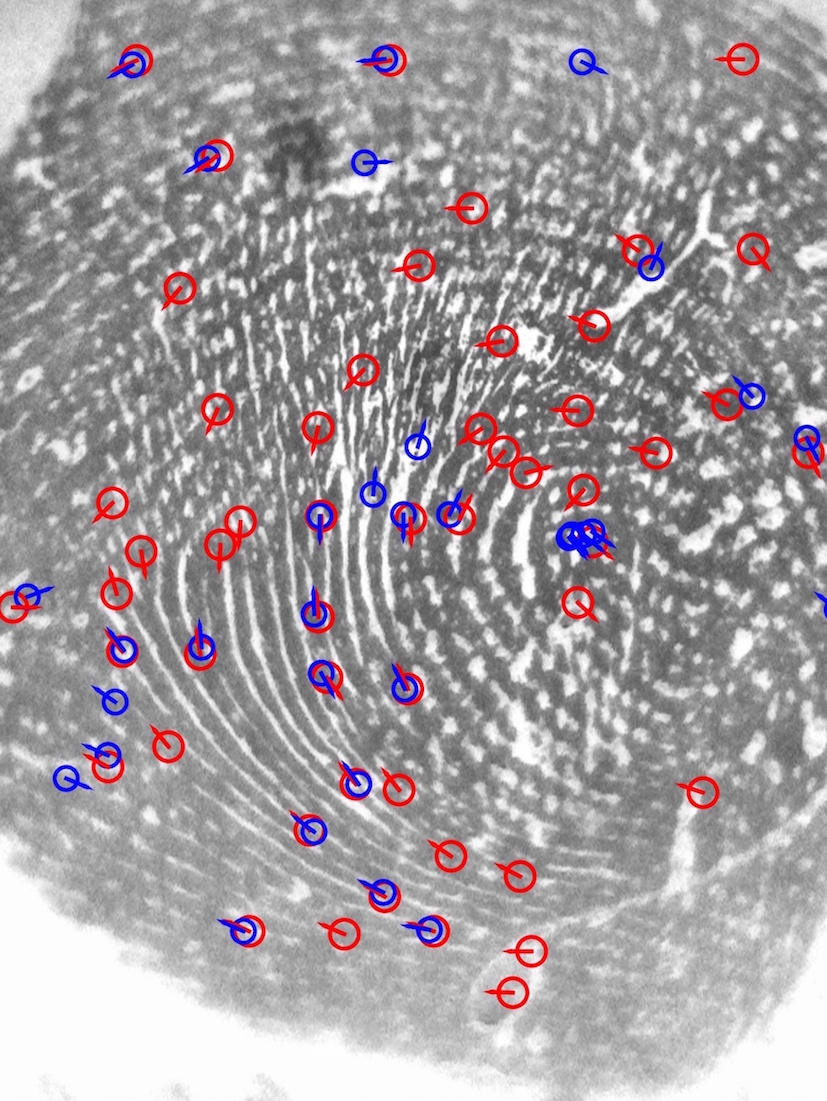}}\hfill
    \subfloat[]{\includegraphics[width=0.19\linewidth]{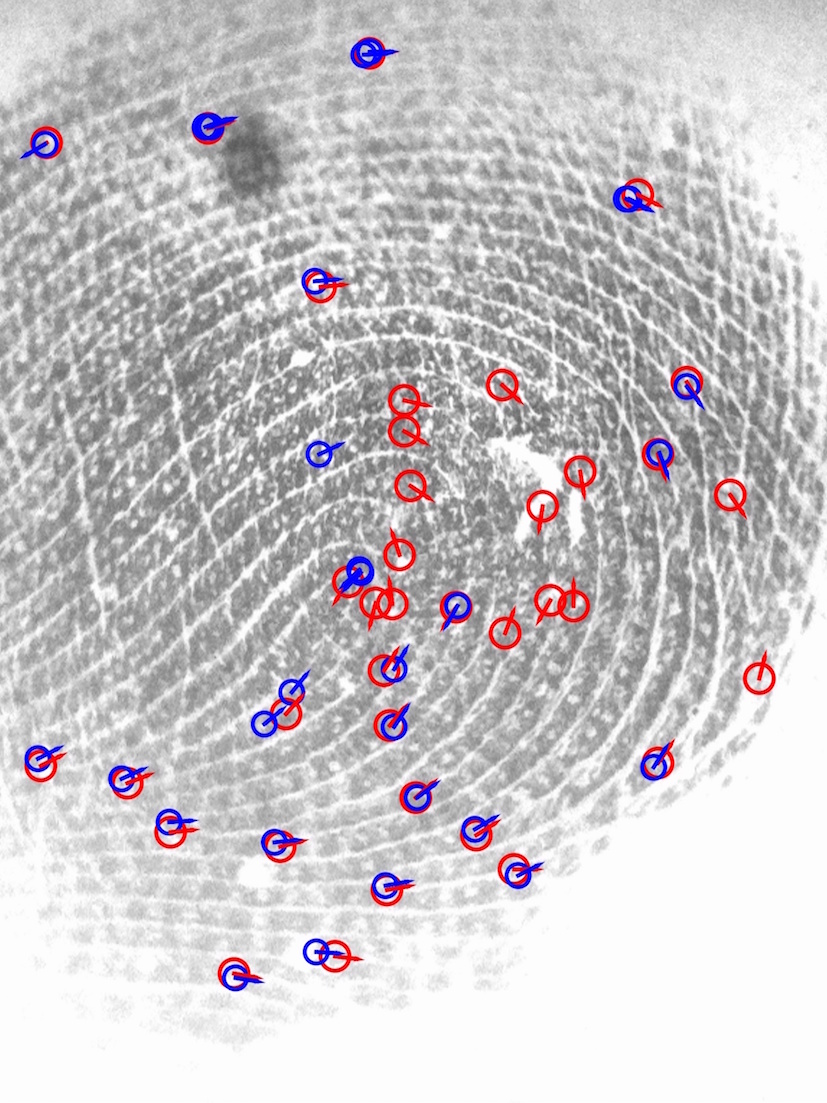}}\\
    \caption{\textbf{Top row:} Verifinger minutiae detections; \textbf{Bottom row:} Minutiae detections from our high-resolution minutiae extractor. Manually marked minutiae are annotated in red. Note that Verifinger detects many of the true minutiae, but also extracts a significant number of spurious minutiae. Our proposed minutiae extractor has slightly lower detection accuracy (of true minutiae) than Verifinger, however, it extracts significantly less spurious minutiae. We further compare the two approaches more quantitatively in our experimental results. }
    \label{fig:markups_eg}
\end{figure*}

The efficacy of our high-resolution minutiae extraction algorithm is shown in Fig.~\ref{fig:markups_eg}. In comparison to Verifinger, our algorithm extracts significantly fewer spurious minutiae, while detecting nearly all of the true minutiae locations. We show in subsequent experiments that this results in a boost in infant fingerprint recognition performance. 

\subsubsection{Minutiae Aging}

After extracting a minutiae set from an infant fingerprint with our high-resolution minutiae extractor, we further process the minutiae set via a minutiae aging model (Fig.~\ref{fig:aging}). The authors in~\cite{growth2} showed that by linearly scaling an infant's fingerprint image, it could be better matched to an older fingerprint impression of the same infant. Note, that although the aging model in~\cite{growth2} was shown to be beneficial for infant recognition, it did not result in desired levels of recognition accuracy due in part to the fact that the infant fingerprint images were captured at 500 ppi.

Rather than scaling an infant's fingerprint \textit{image} as was done in~\cite{growth2}, we directly scale the already extracted \textit{minutiae set}. More formally, given a scale factor $\lambda$ and a minutiae set $T$ of $N$ minutiae, where $T = \{(x_1, y_1, 
\theta_1), ..., (x_N, y_N, \theta_N)\}$, our scaled minutiae set $\hat{T}$ is given by: 

\begin{equation}
\hat{T} = \{(\lambda x_1, \lambda y_1, \theta_1), ...,  (\lambda x_N, \lambda y_N, \theta_N)\}
\end{equation}

\begin{figure*}[!t]
    \centering
    \includegraphics[width=\textwidth]{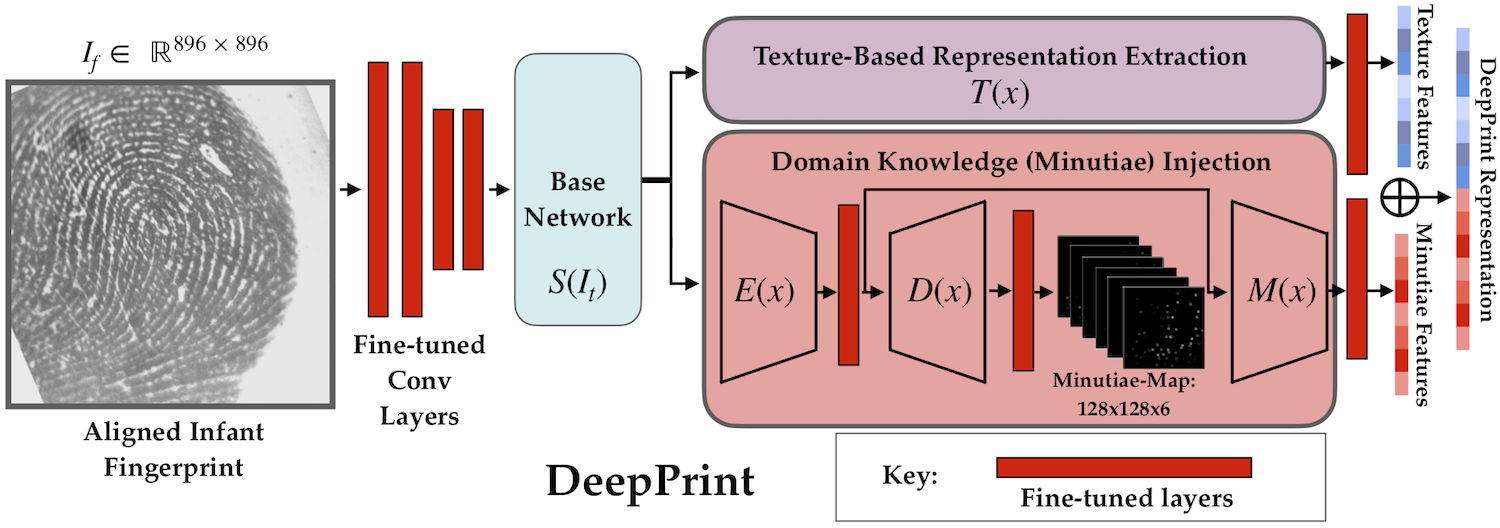}
    \caption{Overview of the Infant-Prints texture matcher. We modify DeepPrint~\cite{engelsma2019learning} to accept 1,900 ppi high resolution infant fingerprint images. The network is pretrained on adult fingerprint images and then fine-tuned (red layers) with the infant dataset collected in~\cite{jain}.}
    \label{fig:overview}
\end{figure*}

To determine the scale factor $\lambda$ at which an infant's fingerprint pattern grows as they age, we select $82$ pairs of our $610$ manually annotated infant fingerprints for which we have longitudinal impressions. The range of the time lapse $\Delta T$ (in weeks) for these $82$ pairs of fingerprints is $12 \leq \Delta T \leq 40$ (mean $\Delta T = 34.3 \pm 10.3$). We then empirically evaluated different scalar factors in increments of $0.05$ such that the minutiae matching accuracy (as computed by Verifinger v10 SDK) on these validation images was maximized. We found that applying a scalar factor of $\lambda = 1.1$ to infant images enrolled at less than 3 months provided the best recognition performance. 

We also tried an adaptive aging model where the scalar factor was dependent upon the enrollment age and the elapsed time, but found no improvement in performance (likely because the majority age group in our experiments is infants enrolled between 2-3 months and recognized 3 months later, where the simple scalar value of $\lambda = 1.1$ suffices). Given similar performance, we kept the simpler static scalar aging model as opposed to the adaptive aging model.

An example of an infant minutiae set $T$ and its corresponding aged minutiae set $\hat{T}$ is shown in Figure~\ref{fig:aging}. In our experiments, we quantitatively demonstrate that this scaling of the enrollment minutiae points provides a boost to our recognition performance. 

\subsubsection{Minutiae Match Score}

After extracting a minutiae set $T$ (via our high-resolution minutiae extractor) and aging $T$ into $\hat{T}$, we compute a minutiae matching score $s_m$ between a probe infant fingerprint and an enrolled infant fingerprint using the Verifinger v10 ISO minutiae matcher. 

\subsection{Texture Matcher}

Similar to latent fingerprints, infant fingerprints are often of poor quality and as such are difficult to accurately extract minutiae from (even with our high resolution minutiae extractor). Therefore, in addition to a minutiae match score, we also incorporate a texture matching score $s_t$ using a state-of-the-art texture fingerprint matcher~\cite{engelsma2019learning}~\footnote{Although DeepPrint also incorporates minutiae domain knowledge into the fixed-length representation, we refer to it as a texture matcher since minutiae points are not explicitly used for matching.}. Engelsma~\textit{et al.}~\cite{engelsma2019learning} proposed a CNN architecture, called DeepPrint, embedded with fingerprint domain knowledge for extracting discriminative fixed-length fingerprint representations. Inspired by the success of DeepPrint to learn additional textural cues that go beyond just minutiae points, we adopt this matcher for infant fingerprint recognition. In particular, we modify the DeepPrint network architecture as follows: (i) the input size of $448 \times 448$ is increased to $1024 \times 1024$ (through the addition of convolutional layers) to support 1,900 ppi images and (ii) the parameters of the added convolutional layers and the last fully connected layer are re-trained on the 1,270 ppi (upsampled to 1,900 ppi) longitudinal infant fingerprints acquired by Jain~\textit{et al.} in~\cite{jain} combined with 610 of our 1,900 ppi images which we set aside for training. In total, we re-train the network with 9,683 infant fingerprint images from 1,814 different thumbs. An overview of our modifications to DeepPrint is shown in Figure~\ref{fig:overview}. 

During the authentication or search stage, the CNN accepts a 1,900 ppi infant fingerprint as input and outputs a 192-dimensional fixed-length representation of the fingerprint. This representation can be compared to previously enrolled representations via the cosine distance between two given representations at 10 million comparisons/second on an Intel i9 processor with 64 GB of RAM. More formally, given an enrollment representation $\mathbf{e} \in \mathbb{R}^{192}$ and a probe representation $\mathbf{p} \in \mathbb{R}^{192}$, a texture matching score $s_t$ is computed as the inner product between $\mathbf{e}$ and $\mathbf{p}$:

\begin{equation}
\label{eq:tscore}
    s_t = \mathbf{e^{T}}\mathbf{p}
\end{equation}

Note, in our preliminary study~\cite{infant_prints}, we also used a deep learning based texture matcher similar to DeepPrint, however, we did not incorporate minutiae domain knowledge into the texture matcher as is done in DeepPrint (shown in Fig.~\ref{fig:overview}). Adopting the strategy of DeepPrint in incorporating minutiae domain knowledge into the deep network further improves the infant recognition performance. We show this quantitatively in the experimental results. 

\subsection{Latent Fingerprint Matcher}

Finally, in addition to a state-of-the-art minutiae matcher (supplemented by our high resolution minutiae extractor) and the fine-tuned texture matcher, we include a state-of-the-art latent fingerprint matcher\footnote{We cannot release the name of the matcher because of a NDA, but it is one of the top performing algorithms in the NIST ELFT evaluation~\cite{elft}.} to the final infant fingerprint recognition algorithm. Before using the latent fingerprint matcher to enroll a template, we first include two preprocessing steps: (i) enhancement, and (ii) aging. These preprocessing steps are further described in the following subsections.

\subsubsection{Enhancement}

\begin{figure}[!h]
    \centering
    \subfloat[]{\includegraphics[width=0.45\linewidth]{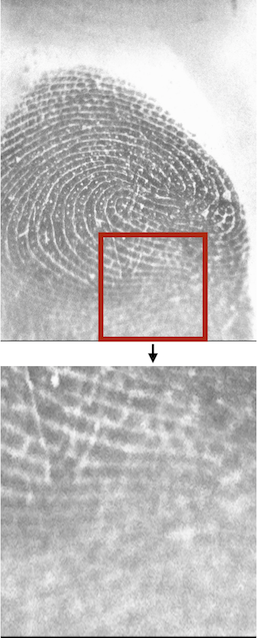}}\hfill
    \subfloat[]{\includegraphics[width=0.45\linewidth]{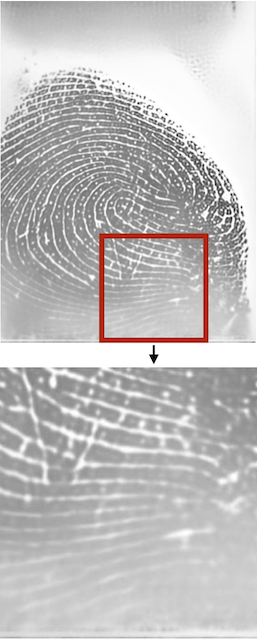}}\\
    \caption{Infant fingerprint (a) before enhancement and (b) after enhancement. The enhanced infant fingerprint (b) has noticeably improved sharpness and clarity throughout the friction ridge pattern when compared to (a).}
    \label{fig:enhancement}
\end{figure}

Due to the low quality of the infant fingerprints (motion blur, wet, dry), we incorporate an enhancement module to improve the sharpness and clarity of the infant friction ridge pattern. In particular, we incorporate a state-of-the-art image super resolution model, Residual Dense Network (RDN)~\cite{sr}. To retrain RDN for infant fingerprint enhancement, we first add random noise (random kernel) to the training dataset (9,683 images from~\cite{jain}), followed by a gaussian blur to simulate various types of noise in the infant fingerprint images. Then, we retrain the RDN network (8x version with a modified stride length) to regress to the clean infant fingerprint images. An example of an infant fingerprint before and after enhancement is shown in Figure~\ref{fig:enhancement}.

\subsubsection{Image Aging}

In a similar manner to the strategy we used to age our extracted minutiae sets, we age the enhanced fingerprint \textit{images} prior to passing them to the latent fingerprint matcher. The COTS latent matcher SDK does not accept a minutiae set and as such, we must directly age the images prior to passing them to the matcher. Therefore, if an infant's fingerprint image is captured at an age of less than 3 months, we resize the image with bicubic interpolation by a scalar factor of $\lambda=1.1$. The scalar factor is the same as that used to scale our minutiae sets. Finally, after enhancement and image aging, we finish the latent preprocessing by resizing all images by a scalar of $0.5$ in order to bring the 1,900 ppi fingerprint images to similar size as the adult fingerprint images the latent matcher is designed to operate on (this same procedure was utilized in~\cite{jain}).

After preprocessing the infant fingerprint images via enhancement and aging, we can enroll the infant images via the latent SDK, and subsequently compute a match score $s_{l}$. 

\subsection{Final Match Score}

Our final match score $s_f$ is a fusion of a minutiae matcher, texture matcher, and latent matcher. In particular, given our minutiae matching score of $s_m$, our texture match score $s_t$ as defined in Equation~\ref{eq:tscore}, and our latent match score $s_l$, our final match score $s_f$ is computed by first normalizing each score (min-max normalization) to a range of $(0, 1)$ and then performing sum score fusion via:

\begin{equation}
\label{eq:fscore}
    s_f = \lambda_m \cdot s_m + \lambda_t \cdot s_t + \lambda_l \cdot s_l
\end{equation} where $\lambda_m$, $\lambda_t$, and $\lambda_l$ are set to $0.6, 0.1, 0.3$ using our validation set of $610$ manually marked fingerprint images in conjunction with a grid search. 

\begin{table*}[t]
\caption{Infant Authentication Accuracy ($0-3$ months at enrollment with 3 month time lapse between enrollment and authentication)}
 \centering
\begin{threeparttable}
\begin{tabular}{c c c c}
 \toprule
 Algorithm & \specialcell{Enrollment Age: 0-1 months \\ (17 subjects) \\ TAR @ FAR=0.1\%, FAR=1.0\%} & \specialcell{Enrollment Age: 1-2 months \\ (36 subjects) \\ TAR @ FAR=0.1\%, FAR=1.0\%} & \specialcell{Enrollment Age: 2-3 months \\ (83 subjects) \\ TAR @ FAR=0.1\%, FAR=1.0\%} \\
 \toprule
 DeepPrint~\cite{engelsma2019learning} & 17.6\%, 29.4\% & 27.8\%, 58.3 & 45.8\%, 68.7\% \\
 \midrule
 Verifinger\tnote{1} & 41.2\%, 58.8\% & 47.2\%, 55.6\% & 79.5\%, 86.7\% \\
 \midrule
 Latent Matcher\tnote{2} & 41.2\%, 47.1\% & 50.0\%, 61.1\% & 84.3\%, 91.6\% \\
 \midrule
 \specialcell{DeepPrint + Verifinger} & 52.9\%, 64.7\% & 55.6\%, 75.0\% & 86.7\%, 89.2\% \\
 \midrule
 \specialcell{DeepPrint + Latent Matcher} & 41.2\%, 58.8\% & 52.8\%, 72.2\% & 85.5\%, 91.6\% \\
 \midrule
 \specialcell{Verifinger + Latent Matcher} & 52.9\%, 64.7\% & 58.3\%, 75.0\% & 91.6\%, 92.8\% \\
 \midrule
 \specialcell{DeepPrint + Verifinger + \\ Latent Matcher} & \textbf{64.7\%, 70.6\%} & \textbf{63.9\%, 83.3\%}  & \textbf{92.8\%, 95.2\%} \\
 \bottomrule
\end{tabular}
\begin{tablenotes}
\item[1] Minutiae are extracted with our high-resolution minutiae extractor, then aged and fed into the Verifinger v10 ISO Matcher.
\item[2] Images are enhanced, aged, and then fed into a state-of-the-art COTS Latent Matcher.
\end{tablenotes}
\end{threeparttable}
\label{table:auth_3_months}
\end{table*}

\begin{table*}[t]
\caption{Infant Search Accuracy ($0-3$ months at enrollment with 3 month time lapse between enrollment and search)}
 \centering
\begin{threeparttable}
\begin{tabular}{c c c c}
 \toprule
 Algorithm & \specialcell{Enrollment Age: 0-1 months \\ (17 subjects) \\ Rank 1, Rank 5} & \specialcell{Enrollment Age: 1-2 months \\ (36 subjects) \\ Rank 1, Rank 5} & \specialcell{Enrollment Age: 2-3 months \\ (83 subjects) \\ Rank 1, Rank 5} \\
  \toprule
 DeepPrint~\cite{engelsma2019learning} & 52.9\%, 58.8\% & 63.9\%, 75.0 & 90.4\%, 92.8\% \\
 \midrule
 Verifinger\tnote{1} & 58.8\%, 64.7\% & 69.4\%, 77.8\% & 90.4\%, 91.6\% \\
 \midrule
 Latent Matcher\tnote{2} & 52.9\%, 58.8\% & 63.9\%, 75.0\% & 90.4\%, 92.8\% \\
 \midrule
 \specialcell{DeepPrint + Verifinger} & 58.8\%, 64.7\% & 69.4\%, 77.8\% & 90.4\%, 91.6\% \\
 \midrule
 \specialcell{DeepPrint + Latent Matcher} & 52.9\%, 58.8\% & 63.9\%, 75.0\% & 90.4\%, 92.8\% \\
 \midrule
 \specialcell{Verifinger + Latent Matcher} & 58.8\%, 58.8\% & 72.2\%, 80.6\% & 90.4\%, 91.6\% \\
 \midrule
 \specialcell{DeepPrint + Verifinger + \\ Latent Matcher} & \textbf{58.8\%, 58.8\%} & \textbf{72.2\%, 77.8\%}  & \textbf{90.4\%, 91.6\%} \\
 \bottomrule
\end{tabular}
\begin{tablenotes}
\item[1] Minutiae are extracted with our high-resolution minutiae extractor, then aged and fed into the Verifinger v10 ISO Matcher.
\item[2] Images are enhanced, aged, and then fed into a state-of-the-art COTS Latent Matcher.
\end{tablenotes}
\end{threeparttable}
\label{table:search_3_months}
\end{table*}

\begin{table*}[t]
\caption{Ablated Infant Authentication Accuracy ($0-3$ months at enrollment with 3 month time lapse between enrollment and authentication)}
 \centering
\begin{threeparttable}
\begin{tabular}{c c c c}
 \toprule
 Algorithm\tnote{\textdagger} & \specialcell{Enrollment Age: 0-1 months \\ (17 subjects) \\ TAR @ FAR=0.1\%, FAR=1.0\%} & \specialcell{Enrollment Age: 1-2 months \\ (36 subjects) \\ TAR @ FAR=0.1\%, FAR=1.0\%} & \specialcell{Enrollment Age: 2-3 months \\ (83 subjects) \\ TAR @ FAR=0.1\%, FAR=1.0\%} \\
  \toprule
 \specialcell{\textit{w/o} High Resolution \\Minutiae Extractor} & 35.3\%, 70.6\% & 63.9\%, 83.3\% & 90.4\%, 95.2\% \\
 \midrule
 \specialcell{\textit{w/o} Aging and \\Enhancement} & 47.1\%, 64.7\% & 50.0\%, 72.2\% & 86.7\%, 92.8\% \\
 \midrule
 \specialcell{\textit{w/o} Finetuning \\DeepPrint} & 58.8\%, 64.7\% & 58.33\%, 69.4\% & 90.4\%, 95.2\% \\
 \midrule
  \specialcell{\textit{w/o} Gender} & 58.8\%, 64.7\% & 52.8\%, 80.6\% & 89.2\%, 94.0\% \\
 \midrule
 \specialcell{\textit{w/o} All\tnote{1}} & 35.3\%, 47.1\% & 44.4\%, 66.7\% & 86.7\%, 92.8\% \\
 \midrule
 \specialcell{\textit{with} All\tnote{2,3}} & \textbf{64.7\%, 70.6\%} & \textbf{63.9\%, 83.3\%}  & \textbf{92.8\%, 95.2\%} \\
 \bottomrule
\end{tabular}
\begin{tablenotes}
\item[1] Algorithm used in our preliminary study~\cite{infant_prints}.
\item[2] Minutiae are extracted with our high-resolution minutiae extractor, then aged and fed into the Verifinger v10 ISO Matcher.
\item[3] Images are enhanced, aged, and then fed into a state-of-the-art COTS Latent Matcher.
\item[\textdagger] Each row removes only the modules mentioned in that row.
\end{tablenotes}
\end{threeparttable}
\label{table:ablation}
\end{table*}

\section{Experimental Results}

In our experimental results, we first show the authentication and search performance for all the infants in our dataset where enrollment occurs during 0-3 months of age, and authentication or search commences 3 months later. We first focus on a 3 month time lapse for the following reasons. (i) Most of our longitudinal data (121 subjects) has a time lapse of 3 months. (ii) Jain \textit{et al.} already show that once infants reach the age of 6 months, they can be enrolled and recognized a year later. In this work, our primary aim is to bridge the gap between 0-3 months (when first time vaccinations commence) and 6 months. If we can effectively recognize the infants enrolled at 2-3 months and authenticated or searched at 5-6 months, we can re-enroll the infants and continue to recognize the infants longitudinally as shown in~\cite{jain}.

We conclude the experiments by showing the authentication and search performance of Infant-Prints when the time lapse between the enrollment and probe images is extended to a year.

\subsection{Experimental Protocol}

To boost the infant recognition performance, we fuse scores from both of the infant's thumbs and also across the multiple impressions captured during the enrollment session and authentication or search session. For example, if we successfully captured 2 fingerprint images of each thumb in the enrollment session and authentication session, we would compute a total of 8 scores using Equation~\ref{eq:fscore}. These 8 scores are then fused using average fusion. 

We also utilize the gender of the infant to further improve the recognition performance. In particular, if two infants have a different gender, we set the matching score to 0.

All imposter scores are computed by comparing impressions from one subject (both thumbs) in a particular session to impressions from another subject (both thumbs) in another session (making sure to only compare impressions if they belong to the same thumb).

\subsection{Infant Authentication}

Table~\ref{table:auth_3_months} shows the authentication performance of the different matchers (as well as the fused matchers) on infants enrolled between the ages of 0-3 months, and authenticated 3 months later. From these results, we observe that none of the individual matchers perform particularly well on any of the age groups when run standalone. However, after fusing the 3 matchers together, we start to get reliable authentication results when the enrollment age is 2-3 months. While the longitudinal authentication results are not yet robust for the age groups of 0-1 months and 1-2 months, we note that vaccinations commence by the age of 3 months. By obtaining promising authentication results at enrollment ages of less than 3 months, we show that fingerprint authentication of infants is indeed a potential solution for providing infants an identity for life.

\subsection{Infant Search}

Table~\ref{table:search_3_months} shows the Rank 1 search accuracy of Infant-Prints on infants enrolled between the ages of 0-3 months, and searched 3 months later. The gallery size for our search experiment includes every infant which was enrolled in our study (315 infants). We acknowledge that this gallery size is small, however, we note that (i) obtaining a large gallery of infants would require significant resources, man-hours, and IRB regulations and approvals, and (ii) in several applications, it is very possible that the gallery size would be of similar size to ours. For example, if the clinic which we collected our data at were to use Infant-Prints, they would only need to manage a gallery of 315 infants, since that is the total number of infants visiting the clinic in a 1 year time period. 

We note from the results of Table~\ref{table:search_3_months} that Infant-Prints is able to enroll infants at an age of 2-3 months, and search them 3 months later with a Rank 1 search accuracy of 90.4\%. While work remains to be done to further improve the performance to say 99\%, we note that this is the first study to show promising longitudinal search performance for infants enrolled at ages as young as 2 months. 

It can also be seen from Table~\ref{table:search_3_months} that each individual matcher is able to obtain the same Rank-1 search performance (for the 2-3 month enrollment group) as the fused matcher. We acknowledge that this can likely be explained by the small gallery size, \textit{i.e.} each individual matcher is sufficient to accurately retrieve the fingerprints from the smaller gallery. Given a larger gallery, it is likely that the fused matcher would be necessary to maintain accurate search performance. Obtaining a large scale infant dataset is an area of future research. 

We also highlight that DeepPrint is able to obtain much higher search performance than authentication performance (Table~\ref{table:auth_3_months} vs. Table~\ref{table:search_3_months}). This can be attributed to DeepPrint often times outputting high imposter scores (creating false accepts and reducing the authentication accuracy, whereas in search high imposters are not as problematic as long as the true mate gives the highest score).

\subsection{Ablations}

To highlight the hardware and algorithmic contributions of Infant-Prints, we show an algorithmic ablation study in Table~\ref{table:ablation}, and a hardware ablation study in Table~\ref{table:ablate_dp}. 

From the algorithmic ablation study (Table~\ref{table:ablation}), it can be seen that every algorithmic improvement (high-resolution minutiae extraction, aging, enhancement, finetuning DeepPrint, and gender meta-data) contributes to the overall best performance shown in the final row. We also note that our algorithm (last row of Table~\ref{table:ablation}) is significantly improved over our previous algorithm (second to last row of Table~\ref{table:ablation}) used in our preliminary study~\cite{infant_prints}.

Finally, we show in our hardware ablation study in Table~\ref{table:ablate_dp} that our contact-based high-resolution (1,900 ppi) fingerprint reader enables higher infant fingerprint authentication performance than a COTS 500 ppi contact-based reader (Digital Personna). We note that there are fewer subjects in Table~\ref{table:ablate_dp} than Table~\ref{table:auth_3_months}. This is because Table~\ref{table:ablate_dp} only considers those subjects which were collected on both the MSU RaspiReader and the Digital Persona reader. The difference in subject counts on the MSU RaspiReader and the Digital Persona reader can be attributed to failure to captures on the Digital Persona (often times the ergonomics of the Digital Persona reader (Fig.~\ref{fig:infant_fpt} (a)) prevented us from imaging the infant's fingerprints before the infant became too distressed). 

\begin{figure}[h]
    \centering
    \subfloat[]{\includegraphics[width=0.95\linewidth]{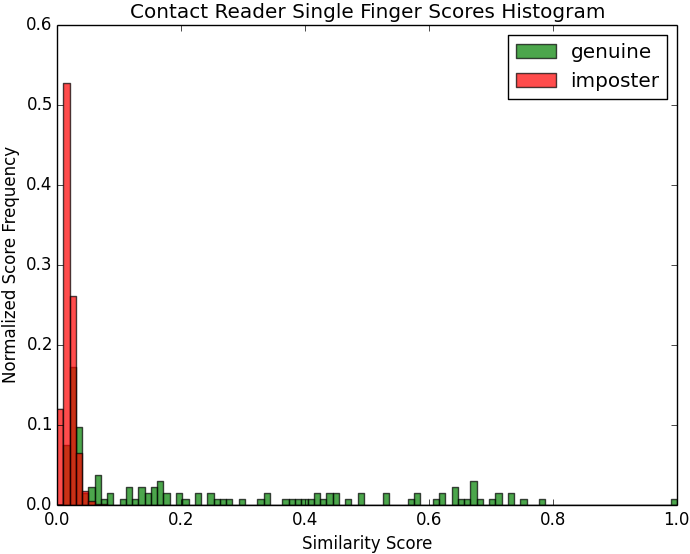}}\hfill
    \subfloat[]{\includegraphics[width=0.95\linewidth]{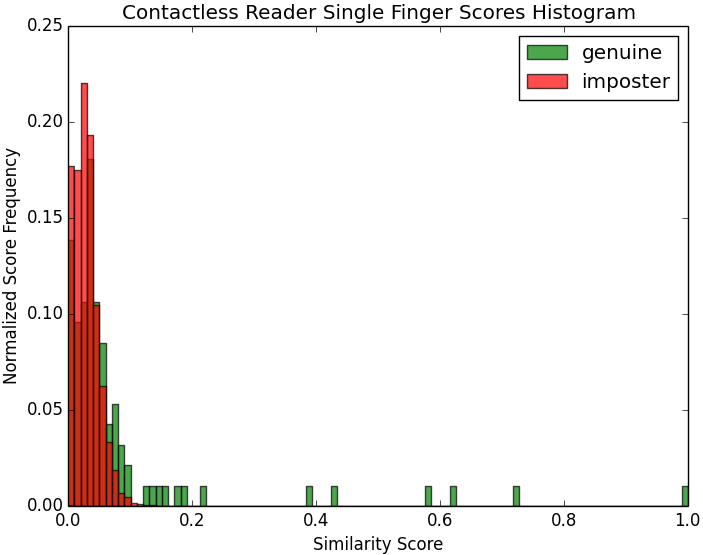}}
    \caption{Score Histograms comparing the contact-based RaspiReader with the contactless RaspiReader. Using a contact-based reader shows much better score separation than the contactless reader.}
    \label{fig:contact_vs_contactless}
\end{figure}

We also show in Figure~\ref{fig:contact_vs_contactless} that the contact-based RaspiReader genuine and imposter scores are much more separated than the contactless-based RaspiReader. We show score histograms (of single finger comparisons) to compare these two readers since we only utilized the contactless reader during our last collection session for a limited number of subjects. Our findings of better separation between the contact fingerprint pairs than the contactless fingerprint pairs \textit{contradict} the study of~\cite{gates} which found that high-resolution, contactless infant fingerprints outperformed high-resolution contact-based infant fingerprints. We found it very difficult to match contactless infant fingerprints since contactless fingerprints have a perspective deformation (certain parts of the finger are further from the camera than others), and the contrast is lower than FTIR fingerprint images. Similar observations about the difficulty of matching contactless fingerprint images have been noted in the literature~\cite{lin2018matching}. 

\begin{table}[t]
\caption{Ablated Fingerprint Reader Authentication Results}
 \centering
\begin{threeparttable}
\begin{tabular}{c c c c}
 \toprule
 Reader & \specialcell{0-1 \\months\tnote{2} \\ (12 subjects)} & \specialcell{1-2 \\months \\ (31 subjects)}  & \specialcell{2-3 \\months \\ (73 subjects)} \\
 \toprule
\specialcell{Digital Persona \\(500 ppi)} & 0\%\tnote{1} & 35.5\% & 52.1\% \\
\midrule
\specialcell{MSU RaspiReader \\ (1,900 ppi)} & 58.3\% & 64.5\% & 93.2\%\tnote{3} \\
 \bottomrule
\end{tabular}
\begin{tablenotes}
\item[1] TAR @ FAR = 0.1\% after a time lapse of 3 months from enrollment age.
\item[2] Indicates enrollment ages (authentication occurs 3 months later).
\item[3] Differs from Table~\ref{table:auth_3_months} because of a different number of subjects.
\end{tablenotes}
\end{threeparttable}
\label{table:ablate_dp}
\end{table}

\begin{figure}[h]
    \centering
    \subfloat[]{\includegraphics[width=0.45\linewidth]{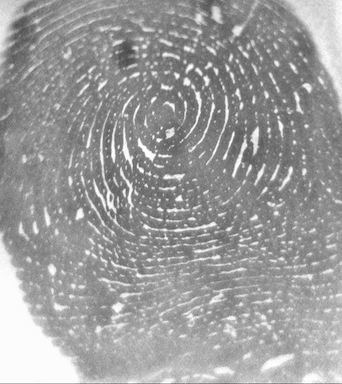}}\hfill
    \subfloat[]{\includegraphics[width=0.45\linewidth]{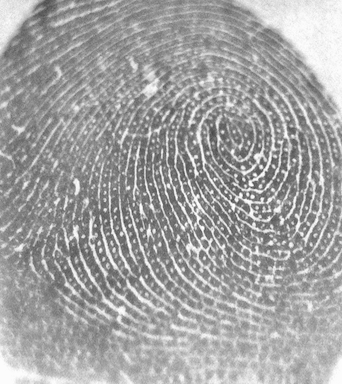}}\hfill
    \subfloat[]{\includegraphics[width=0.45\linewidth]{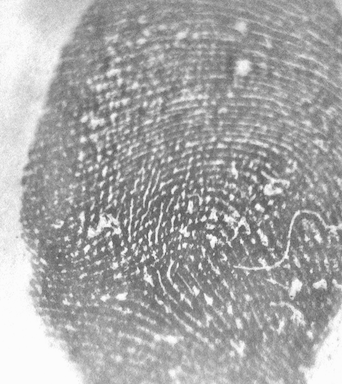}}\hfill
    \subfloat[]{\includegraphics[width=0.45\linewidth]{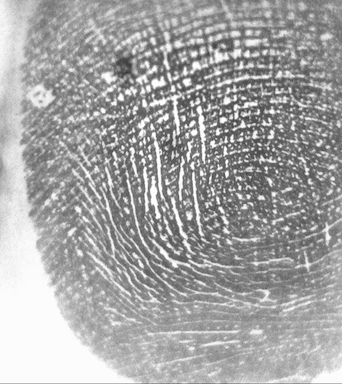}}\\
    \caption{Example Infant-Prints failure cases. (a, b) Example of a False Accept due to the similar friction ridge patterns, and the moisture in the enrollment image (a). (c, d) Example of a False Reject due to the motion blur of the uncooperative infant (d). These images highlight several of the challenges in infant fingerprint recognition.}
    \label{fig:failures}
\end{figure}

Example of failure cases (False Accept, False Reject) are shown in Fig.~\ref{fig:failures}. These images highlight the difficulty and challenges of doing accurate infant fingerprint recognition over time (moisture, distortion, small inter-ridge spacing, fingerprint aging). 

\subsection{Longitudinal Recognition}

As a final study, we show the longitudinal search accuracy (Table~\ref{table:long_search}) and authentication accuracy (Table~\ref{table:long_auth}) for infants enrolled at 2-3 months. For this experiment, we selected 20 infants from our total of 315 which were present in all 4 sessions of the data collection and were 2-3 months of age at the first time enrollment (since our earlier studies showed that 2-3 months is the age at which recognition first becomes feasible).   Although we have more subjects at individual time lapses, we chose the 20 infants which were present in all 4 sessions so that we can observe the impact that time has on the recognition performance whilst fixing the subjects used in the experiments. 

Tables~\ref{table:long_search} and~\ref{table:long_auth} show that the authentication and search performance stays relatively stable over time. In particular, from 3 months of elapsed time to 9 months of elapsed time, only one infant drops off from being properly searched or authenticated. From 9 months to 12 months, the search accuracy remains unchanged, while only one fewer infant is unable to be authenticated.

Notably, these are the first results to show that it is possible to enroll infants at 2 months old and authenticate them or search them a year later with relatively high accuracy. This highlights the applicability of fingerprints to address the challenges of this paper. Namely, \textit{can we recognize an infant from their fingerprints in order to better facilitate accurate and fast delivery of vaccinations and nutritional supplements to infants in need}.

\begin{table}[!h]
\caption{Longitudinal Search Results}
 \centering
\begin{threeparttable}
\begin{tabular}{c c c}
 \toprule
 \specialcell{Time Lapse: 3 months} & \specialcell{Time Lapse: 9 months} & \specialcell{Time Lapse: 12 months} \\
  \toprule
95\%\tnote{1, 2} & 90\% & 90\% \\
 \bottomrule
\end{tabular}
\begin{tablenotes}
\item[1] Reporting Rank 1 Search Accuracy (Gallery of 315 Infants)
\item[2] Differs from Table~\ref{table:search_3_months} because of a different number of subjects.
\end{tablenotes}
\end{threeparttable}
\label{table:long_search}
\end{table}

\begin{table}[!h]
\caption{Longitudinal Authentication Results}
 \centering
\begin{threeparttable}
\begin{tabular}{c c c}
 \toprule
 \specialcell{Time Lapse: 3 months} & \specialcell{Time Lapse: 9 months} & \specialcell{Time Lapse: 12 months} \\
  \toprule
95\%\tnote{1,2} & 90\% & 85\% \\
 \bottomrule
\end{tabular}
\begin{tablenotes}
\item[1] Reporting TAR @ FAR = 0.1\%
\item[2] Differs from Table~\ref{table:auth_3_months} because of a different number of subjects.
\end{tablenotes}
\end{threeparttable}
\label{table:long_auth}
\end{table}

\section{Conclusion}

A plethora of infants around the world continue to suffer and die from vaccine related diseases and malnutrition. A major obstacle standing in the way of delivering the vaccinations and nutrition needed to the infants most in need is the means to quickly and accurately identity or authenticate an infant at the point of care. To address this challenge, we proposed Infant-Prints, and end-to-end infant fingerprint recognition system. We have shown that Infant-Prints is capable of enrolling infants as young as 2 months of age, and recognizing them an entire year later. This is the first ever study to show the feasibility of recognizing infants enrolled this young after this much time gap. It is our hope that this feasibility study and Infant-Prints motivate a strong push in the direction of fingerprint based infant fingerprint recognition systems which can be used to alleviate infant suffering around the world. In doing so, we believe that this work will make a major dent in Goal \#3 of the United Nations Sustainable Development Goals, namely, \textit{``Ensuring healthy lives and promoting well-being for all, \textbf{at all ages.}"}

{\small
\bibliographystyle{unsrt}
\bibliography{egbib}
}


%



%

\begin{IEEEbiography}[{\includegraphics[width=1in,height=1.25in,clip,keepaspectratio]{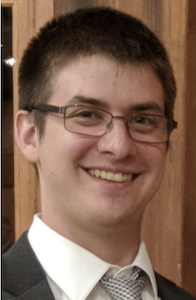}}]{Joshua J. Engelsma}
graduated magna cum
laude with a B.S. degree in computer science
from Grand Valley State University, Allendale,
Michigan, in 2016. He is currently working towards a PhD degree in the Department of
Computer Science and Engineering at Michigan
State University. His research interests include
pattern recognition, computer vision, and image
processing with applications in biometrics. He
won the best paper award at the 2019 IEEE
International Conference on Biometrics (ICB), and the 2020 Michigan State University College of Engineering Fitch Beach Award.
\end{IEEEbiography}


\begin{IEEEbiography}[{\includegraphics[width=1in,height=1.25in,clip,keepaspectratio]{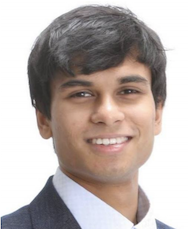}}]{Debayan Deb}
received his B.S. degree in computer
science from Michigan State University, East Lansing, Michigan, in 2016. He is currently working towards a PhD degree in the Department of Computer
Science and Engineering at Michigan State University, East Lansing, Michigan. His research interests
include pattern recognition, computer vision, and
machine learning with applications in biometrics.
\end{IEEEbiography}

\begin{IEEEbiography}[{\includegraphics[width=1in,height=1.25in,clip,keepaspectratio]{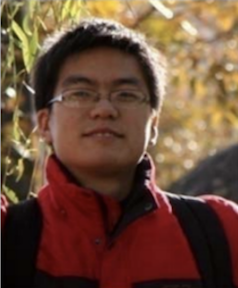}}]{Kai Cao}
received the Ph.D. degree from the
Key Laboratory of Complex Systems and Intelligence Science, Institute of Automation, Chinese Academy of Sciences, Beijing, China, in
2010. He was a Post Doctoral Fellow in the Department of Computer Science \& Engineering,
Michigan State University. He is now a Senior Research Scientist at Goodix in San Diego, CA. His research interests
include biometric recognition, computer vision, image processing
and machine learning.
\end{IEEEbiography}

\begin{IEEEbiography}[{\includegraphics[width=1in,height=1.25in,clip,keepaspectratio]{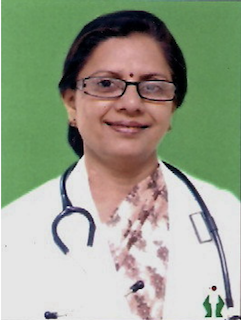}}]{Anjoo Bhatnagar}
has more than 30 years experience in the field of Paediatrics. She is currently Paediatrician
cum Neonatologist at the Saran Ashram Hospital, Dayalbagh at Agra. She has received Presidents Medal during
MBBS for standing first in order of merit in the
university. She is Life member, Indian Academy of
Pediatrics and Founder President of National Neonatology Forum, Faridabad.
She is national trainer in NRP (neonatal resuscitation programme) and has
been selected as nodal person by Govt. of India and UNICEF for SCNU
(special care neonatal unit) project development of NRHM (national rural
health mission) and as trainer of Accredited Social Health Activist (ASHA)
to reduce neonatal mortality in India. Her research interests include child
development and fetal and neonatal consciousness.
\end{IEEEbiography}

\begin{IEEEbiography}[{\includegraphics[width=1in,height=1.25in,clip,keepaspectratio]{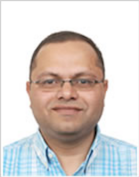}}]{Prem S. Sudhish}
is with the Dayalbagh Educational Institute. His research interests are in frugal innovation, identity, security and machine consciousness. He has also been a volunteer consultant to the Government of India on national projects and works towards bringing cutting edge research and development in technology to benefit the marginalized and underprivileged in the society. Dr. Sudhish has a BS in Electrical Engineering with distinction from Dayalbagh Educational Institute, MS from Stanford University and PhD through a collaborative arrangement between Dayalbagh Educational Institute and Michigan State University. He is also the coordinator for the international agreement for academic cooperation between Michigan State University and Dayalbagh Educational Institute.
\end{IEEEbiography}

\begin{IEEEbiography}[{\includegraphics[width=1in,height=1.25in,clip,keepaspectratio]{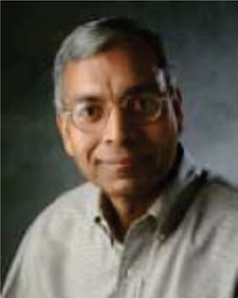}}]{Anil K. Jain}
is a University distinguished professor
in the Department of Computer Science and Engineering at Michigan State University. His research
interests include pattern recognition and biometric
authentication. He served as the editor-in-chief of the
IEEE Transactions on Pattern Analysis and Machine
Intelligence and was a member of the United States
Defense Science Board. He has received Fulbright,
Guggenheim, Alexander von Humboldt, and IAPR
King Sun Fu awards. He is a member of the National
Academy of Engineering and foreign fellow of the
Indian National Academy of Engineering and Chinese Academy of Sciences.
\end{IEEEbiography}




\end{document}